\documentclass[review]{elsarticle}

\usepackage{lineno,hyperref}
\modulolinenumbers[5]

\usepackage{times}
\usepackage{epsfig}
\usepackage{graphicx}
\usepackage{amsmath}
\usepackage{amssymb}

\usepackage{textcomp}
\usepackage{bbding}

\usepackage{tabularx}
\usepackage{caption2}

\usepackage{multirow}
\usepackage[T2A,T1]{fontenc}

\usepackage{subfigure}

\usepackage{bbm}

{\raise0.5ex\hbox{$#1$}\! \left/ \! \lower0.5ex\hbox{$#2$}\right.}

\usepackage{color}

\usepackage[para,online,flushleft]{threeparttable}

\usepackage[noend]{algpseudocode}

\usepackage{algorithmicx,algorithm}
\algrenewcommand\algorithmicrequire{\textbf{Input:}}
\algrenewcommand\algorithmicensure{\textbf{Output:}}

\journal{Neurocomputing}

\begin{document}

\begin{frontmatter}

\title{SA-Net: A deep spectral analysis 
	network for image clustering}

\author{Jinghua Wang and Jianmin Jiang \corref{mycorrespondingauthor}}
\cortext[mycorrespondingauthor]{Corresponding author}
\address{Research Institute for Future Media Computing, College of Computer Science \& Software Engineering, and Guangdong Laboratory of Artificial Intelligence \& Digital Economy (SZ), Shenzhen University, China
	\\ 
wangjh2012@foxmail.com; jianmin.jiang@szu.edu.cn}

\begin{abstract}
Although supervised deep representation learning has attracted enormous attentions across areas of pattern recognition and computer vision,
little progress has been made towards unsupervised deep representation learning for image clustering.  
In this paper, we propose a deep spectral analysis network for unsupervised representation learning and image clustering.
While spectral analysis is established with solid theoretical foundations and has been widely applied to unsupervised data mining, its essential weakness lies in the fact that it is difficult to construct a proper affinity matrix and determine the involving Laplacian matrix for a given dataset.
In this paper, we propose a SA-Net to overcome these weaknesses and achieve improved image clustering by extending the spectral analysis procedure into a deep learning framework with multiple layers. The SA-Net has the capability to learn deep representations and reveal deep correlations among data samples. Compared with the existing spectral analysis, the SA-Net achieves two advantages: (i) Given the fact that one spectral analysis procedure can only deal with one subset of the  given dataset, our proposed SA-Net elegantly integrates multiple parallel and consecutive spectral analysis procedures 
together to enable interactive learning across different units towards a coordinated  clustering model; (ii) Our SA-Net can identify the local similarities among different images at patch level and hence achieves a higher level of robustness against occlusions.
Extensive experiments on a number of popular datasets support that our proposed SA-Net outperforms $ 11 $ benchmarks  across a number of image clustering applications.
\end{abstract}


\end{frontmatter}


\section{Introduction}

As clustering is one of the most fundamental tasks in machine learning and data mining \cite{XU-survey-tnn2005,Yang2010Image,Trigeorgis2014A}, its main goal is to reveal the meaningful structure of a dataset by categorizing the data samples into a number of clusters, where similar samples are grouped together. 
It is extensively studied and a large number of methods have been reported, including
subspace clustering \cite{CHEN2018177-neuro}, partitional clustering \cite{ Wu-trans2015-fuzzy-anew}, hierarchical clustering \cite{Liu-2017pami-hierarchical,Zhang2013Agglomerative}, and density-based clustering \cite{TONG20182355-neurocomputing}.
Over the past decades, clustering has found a wide range of applications, such as  video retrieval \cite{CAI2018316-neurocomputing},  text analysis \cite{LEE2018210-neuro}, as well as large scale data analysis \cite{ZHAO2018227-neurocomputing,Chen2011Large}.

Spectral analysis is one of the most promising clustering methods \cite{Ng01onspectral,Li2015Scalable}, and has been successfully applied in  various computer vision tasks \cite{Nie2011Spectral,Liu2017Spectral}. Spectral analysis first derives a Laplacian matrix from the pairwise similarities among the data samples, and then embeds the data samples into an eigenspace of the Laplacian matrix, before the $ k $-means is applied to complete the final categorization of all the data samples.
Theoretically, the numerical embeddings or the spectral features of the data samples are taken as the relaxation of binary cluster labels \cite{Yu_Shi_multiclass-iccv-2003}. Thus, these spectral features can improve both the intra-cluster compactness and the inter-cluster separability. Spectral analysis has three appealing properties, including (i) it can produce the embeddings analytically via an eigen-decomposition procedure;
(ii) it has solid interpretations and can be derived from the theory of random walk, where the diffusion distance between a pair of data samples is equal to the distance between their embeddings \cite{Nadler:2005:DMS:2976248.2976368}; (iii) spectral analysis is effective in revealing the non-convex data structure \cite{Kang-arxiv-unified-2017}.

In general, spectral analysis has two unsolved problems.
The first problem is the fact that it is still unclear how the affinity graph can influence the clustering performance. 
To construct the affinity graph, there exist three popular strategies, including, $ k $-nearest-neighborhood, $ \epsilon $-nearest-neighborhood, and fully connected graph.
While each of these three strategies has its own advantages and disadvantages, how to choose a specific strategy and how to determine its optimal parameter still remains to be an open issue.
The second problem is that no agreement has ever been reached in the choice of Laplacian matrix for eigenvector decomposition. Both of the two popular Laplacian matrices, i.e. symmetric normalized Laplacian matrix \cite{Ng01onspectral} and left normalized Laplacian matrix \cite{Shi2000-pami-Normalized,vonLuxburg2007-a-tutorial-on-spectral-clustering},  have their own advantages and disadvantages.

Being the input of clustering, representations of data samples are also important for achieving good performances.
Out of the popularity of deep learning, an increasing number of researchers use convolutional neural network (CNN) to learn deep representations that are feasible for clustering \cite{Dizaji-iccv2017-deepClustering,Li2017DiscriminativelyBI,Shaham-iclr-2018SpectralNetSC,Tian:2014:Learning-deep,Xie:icml2016:Unsupevisd-deep-learning,Yang2016Towards,Yang-cvpr2016-joint,wangdeepgmm}. 
Compared with low-level or handcrafted representations, the deep representations show overwhelming strengths in dealing with complicated data sample distributions \cite{Tian:2014:Learning-deep,Aljalbout2018ClusteringWD,chan-2015-tip-pcanet}.

Motivated by the significant success of deep learning, we extend the spectral analysis into multiple layers and propose a new spectral analysis network (SA-Net).
Our SA-Net  learns deep representations based on both parallel and consecutive spectral analysis procedures and shows its strength in various image clustering tasks. The proposed SA-Net consists of four different types of layers, i.e. spectral analysis layer, pooling layer, binarization layer, and coding layer. 
While parallel spectral analysis procedures reveal the intrinsic structure of differently distributed data samples, the consecutive  spectral analysis procedures inside the SA-Net learn deep features to further improve the clustering friendliness of spectral features.
By taking the image patches as the input, a spectral analysis layer learns a patch-level representation space  in such a way that similar patches are made close to each other and  dissimilar ones are made far away from each other.
This procedure can implicitly associate similar patches across different images and  identify the local similarity among them. While the spectral analysis layer stacks the patch representations to produce the  representation of an image for further processing by other layers, the pooling layer reduces the size of the feature by summarization, the binarization layer binarizes the spectral  features, and the coding layer transforms the binary features into numerical feature maps.

Compared with the existing spectral analysis techniques applied to clustering, our SA-Net has the following three  advantages. 
\begin{itemize} 
	\item  While the existing approaches are dominated by one single spectral analysis procedure to learn clustering-friendly features,  our SA-Net stacks multiple spectral analysis procedures in both parallel and consecutive manners to identify the best possible features for data samples across various distribution models.

	\item Our proposed SA-Net elegantly integrates three types of affinity graphs as well as two different normalized Laplacian matrices rather than relying on a single empirically determined spectral analysis procedure. 
	In this way, different spectral analysis procedures can collaborate together in dealing with different data sample distributions.
	Thus, our network can  achieve enhanced clustering performances in dealing with the variety of input datasets.
	\item While existing spectral analysis can only assess the similarity between image pairs holistically, our proposed SA-Net can reveal the local similarity at patch level via learning with multiple and multi-type layers. As a result, the proposed method is more robust in identifying local similarities among images, especially against occlusions.
\end{itemize}

The rest of this paper is organized as follows. Section \ref{sec-related-work} reviews the existing spectral  analysis clustering  and deep representation learning, which are related to our work. Section \ref{sec:the-proposed-method} presents the details of the proposed spectral analysis network (SA-Net). Section  \ref{sec:experiment} reports the experiments and finally Section \ref{sec:conclusion} provides concluding remarks.

\section{Related Work}
\label{sec-related-work}

\subsection{Spectral Analysis Clustering}
\label{subsec-spectral-clustering}

Given a dataset of $ N $ samples, i.e. $ I=\{I_1,I_2,\cdots,I_N \} $, a  clustering task aims to  partition it into $ k $ clusters.
To achieve this, spectral analysis methods first build a non-directed graph $ G=\{I,W\}$, with $ W \in R^{N\times N} $ as the affinity matrix. In the graph $ G $, each node $ I_i (1 \leq i \leq N)$ corresponds to a data sample and the element $ w_{ij} (1 \leq i,j \leq N)$ represents the affinity between a pair of nodes $ I_i $ and $ I_j $.
Spectral analysis  partitions the graph $ G $ into a number of subgraphs based on a graph cut criterion \cite{Shi2000-pami-Normalized}, and thus produces a set of clusters. 
Mathematically, spectral clustering solves the following minimization problem:
\begin{equation}
\label{eq:binary-assignment-matrix}
\min_{\hat{Y}}tr(\hat{Y}^T W \hat{Y})
\end{equation}
where the binary assignment matrix  $ \hat{Y}\in \{0,1\}^{N\times k} $ satisfies  $ \hat{Y}  \mathbbm{1}_k=\mathbbm{1}_N   $, i.e. each sample belongs to one and only one cluster. The element $ \hat{y}_{ic}=1 $ if and only if the data sample $ I_i $ is assigned to the $ c $th cluster.

It is known that the problem in Eq. (\ref{eq:binary-assignment-matrix}) is NP-hard.  In order to solve this problem numerically, researchers relax it by the spectral graph theory \cite{chung-spectral-graph1997}. By allowing the assignment matrix to have continuous values, we obtain the following relaxation for Eq. (\ref{eq:binary-assignment-matrix}): 
\begin{equation}
\label{eq:nonbinary-assignment-matrix}
\min_{{Y}}tr({Y}^T W {Y}) \quad s.t. \quad Y^{T}Y=E_k
\end{equation}
where $ Y\in R^{N\times k} $ is the relaxed continuous clustering assignment matrix with orthogonal constraint and  $ E_k \in R^{k\times k} $ is the identity matrix. 
Based on a normalized cut criterion, the spectral clustering can also be formulated as \cite{Yu_Shi_multiclass-iccv-2003}:
\begin{equation}
\label{eq-normalized-graph-formulation}
\max_{Y^TDY=E_k} tr(Y^TWY)
\end{equation} 
where $ Y=\hat{Y}(\hat{Y}^TD\hat{Y})^{-1/2} $ and $ D \in R^{N\times N} $ is a diagonal matrix with $ D_{ii}=\sum_{j=1}^{N} w_{ij} $.
We can also formulate spectral clustering as \cite{Ng01onspectral}
\begin{equation}
\label{eq-normalized-graph-formulation-laplacian}
\min_{Y^TDY=E_k} tr(Y^TLY)
\end{equation} 
where $ L=D-W $ is the Laplacian matrix.
With the solution $ Y $ of Eq. (\ref{eq:nonbinary-assignment-matrix}), (\ref{eq-normalized-graph-formulation}), or (\ref{eq-normalized-graph-formulation-laplacian}), we can obtain the final cluster labels by simply conducting an additional $ k $-means procedure.
For the convenience of description, we  consider $ y_i $ as the  spectral feature of the data sample $ I_i (1 \leq i \leq N)$.

\begin{algorithm}
	\caption{Spectral Clustering}
	\label{alg:spectral-clustering}
	\begin{algorithmic}[1]
		\Require data points $ I=\{I_1,I_2,\cdots,I_N\} $ and number of clusters $ k $;
		\Ensure $ k $ clusters
		\State  Construct an  affinity matrix $ W \in R^{N \times N} $ between data points, where $ w_{ij} $ measures the similarity between $ I_i $ and $ I_j $;
		\State Compute the Laplacian matrix $ L=D-W $, where $ D \in R^{N\times N} $ is the degree matrix  with  $ d_{ii}=\sum_{j=1}^{N} w_{ij}$;		
		\State Compute the  $ k $ eigenvectors $ q_i (1 \leq i \leq k) $ for Laplacian matrix  $ L $ , corresponding to the $ k $ smallest eigenvalues, and denote them by: $ Q=[q_i, q_2, \cdots, q_k] \in R^{N \times k}$;		
		\State For $ 1\leq i \leq N $, let $ y_i \in R^k $ be the $ i $th row of the matrix $ Q $, and apply $ k $-means to cluster the points $ y_i (1\leq i \leq N) $ to obtain the $ k $ clusters: $ Cluster_j (1\leq j \leq k) $.		
	\end{algorithmic}
\end{algorithm}

A typical spectral analysis clustering procedure is shown in Algorithm \ref{alg:spectral-clustering}, which mainly  consists of four steps, i.e. affinity matrix construction, Laplacian matrix computation, matrix eigen-decomposition, and $ k $-means clustering. 
While spectral analysis clustering has the advantage in revealing the intrinsic data distribution \cite{Ng01onspectral,vonLuxburg2007-a-tutorial-on-spectral-clustering}, it also has a number of deficiencies, compared with other clustering approaches, which can be highlighted as follows.

Firstly, there is little theoretical analysis that could lead us to a proper affinity matrix $ W $ for a given dataset, although it is extensively studied \cite{Zelnik-Manor:2004:Self-tuning,zhu-cvpr2014-constructing}.  
While  three different similarity measurements are  popularly used to construct the affinity matrix, including  $ k $-nearest-neighborhood, $ \epsilon $-nearest-neighborhood, and the fully connected graph, 
each of these three affinity matrices can only deal with some but not all types of data sets.
While the $ k $-nearest-neighborhood strategy might break a connected component into several components, the $ \epsilon $-nearest-neighborhood strategy can not handle datasets with varying densities, and the fully connected affinity method  suffers from high  computational complexity. 
In addition, the clustering results are also sensitive to the parameter variations of these similarity measurements.

Secondly, the relevant research communities have not reached consensus on how to choose between different Laplacian matrices. 
The unnormalized Laplacian matrix  $ L $ has two popular extensions, including symmetric normalization and left normalization, and details of these two normalization matrices are described in the following two equations:
\begin{equation}
\label{eq:symmeteric-Laplacian}
L_{sym}=D^{-1/2}LD^{-1/2}=E_N-D^{-1/2}WD^{-1/2}
\end{equation}
\begin{equation}
\label{eq:left-normalized-matrix}
L_{rw}=D^{-1}L=E_N-D^{-1}W
\end{equation}
where $ E_N \in R^{N\times N} $ is the identity matrix.
Both the  normalized matrices are positive semi-definite and thus have $ N $ real-valued eigenvalues. Let $ v_{rw} $ be an eigenvector of $ L_{rw} $ corresponding to eigenvalue $ \lambda_{rw} $, i.e. $ L_{rw} v_{rw}=\lambda_{rw} v_{rw} $. Then, we have the following  equations to describe their relationships.
\begin{equation}
\begin{split}
& E_Nv_{rw}-D^{-1}Wv_{rw}=\lambda_{rw} v_{rw}  \Leftrightarrow
\\
&v_{rw}-D^{-1/2}D^{-1/2}WD^{-1/2}D^{1/2}v_{rw}=\lambda_{rw} v_{rw}
\Leftrightarrow
\\
&D^{1/2}v_{rw}- (D^{-1/2}WD^{-1/2})D^{1/2}v_{rw}=\lambda_{rw} D^{1/2}v_{rw} \Leftrightarrow 
\\
&
L_{sym}(D^{1/2}v_{rw})=\lambda_{rw} (D^{1/2}v_{rw})
\end{split}
\end{equation}
Thus,  $ D^{1/2}v_{rw} $ is the eigenvector of $ L_{sym} $ corresponding to the eigenvalue $ \lambda_{rw} $. This means that $ L_{sym} $ and $ L_{rw} $ have the same set of eigenvalues and their eigenvectors differ by a scaling of $ D^{1/2} $.
While Ng \cite{Ng01onspectral} adopted the symmetrically normalized Laplacian matrix and claimed superior performances, Shi \cite{Shi2000-pami-Normalized} and Luxburg \cite{vonLuxburg2007-a-tutorial-on-spectral-clustering} recommended the left normalized matrix. Both normalizations have their individual advantages and disadvantages.

Thirdly, spectral clustering is computationally expensive. In general, Algorithm \ref{alg:spectral-clustering} has the computational complexity of $ O(N^2) $  in terms of space and $ O(N^3) $ in terms of time, and thus many efforts have been reported to reduce the computational complexity. 
Dhillon et al. \cite{Dhillon:2007-weighted-graph} eliminate the need for eigen-decomposition by proposing a multilevel algorithm to optimize weighted graph cuts.
Yan et al. \cite{Yan:2009-fast-approximate} propose to conduct spectral clustering based on the representative centroids. In a similar manner, Zhang et al. \cite{Zhang-icml-2008-improved} minimize the quantization error of samples and thus improve Nystrom spectral clustering. 
Based on the Nystrom method, Fowlkes et al. \cite{Fowlkes-tpami2004-spectral-grouping}  approximate the affinity matrix using a subset of data samples. 
Wang et al. \cite{wang-icde-2015-multilevel} select a subset of data points based on data-dependent nonuniform probability distribution and use them to construct a low-rank approximation for the affinity matrix.
In 2017, Han and Filippone \cite{Han-ijcnn2017-miniBatch} propose to recover the Laplacian spectrum via mini-batch-based stochastic gradient optimization on Stiefel manifolds. In contrast, our proposed approach is computationally efficient and converges to critical points, even with a data set as large as $ 580K $, providing the potential of allowing us to conduct spectral analysis on large datasets.

\subsection{Deep Representation Learning for Clustering}
\label{subsec-deepRepresentation-learning}

To achieve satisfactory clustering performances,  data representations are also of vital importance in addition to the clustering techniques. 
Along with the popularity of deep learning \cite{wang-tsvt}, existing research has increasingly focused on deep representations, leading to significant improvement of clustering performances. 
Based on the reconstruction task, Hinton and  Salakhutdinov propose an autoencoder to learn deep representations \cite{Hinton-Science2006-reducing}. 
Tian et al. \cite{Tian:2014:Learning-deep}  correlate spectral clustering with autoencoder, and thus put forward a so-called  sparse autoencoder for deep representation learning. Chen \cite{Chen15a-arxiv-deeplearningwith} takes the nonparametric maximum margin clustering results as the supervision information and  learns the deep data representations using a DBN (deep belief network). 
As a recurrent framework for agglomerative clustering, JULE \cite{Yang-cvpr2016-joint} integrates CNN-based representation learning with the cluster assignment learning, and through such an integration, these two learning procedures can boost each other.
Both DCN \cite{Yang2016Towards} and DBC \cite{Li2017DiscriminativelyBI}  propose  a well-designed objective function in order to learn deep representations which are suitable for $ k $-means clustering. 
DEC \cite{Xie:icml2016:Unsupevisd-deep-learning} proposes soft assignments for  data samples based on the representation distribution and refines them iteratively. 
DEPICT \cite{Dizaji-iccv2017-deepClustering} designs a new network structure by stacking a soft-max layer on top of a multilayer autoencoder and trains it by minimizing relative entropy. Shaham et al. \cite{Shaham-iclr-2018SpectralNetSC} train a SpectralNet that can learn the data embeddings as well as the cluster assignments at the same time. Such spectralNet can deal with out-of-sample problem as well as large data set. Recently, Aljalbout et al. \cite{Aljalbout2018ClusteringWD} present a systematic taxonomy for clustering with deep learning.

All of the above mentioned methods learn deep representations based on convolutional neural networks.
In this work, we extend \cite{wang-icme2019} and propose a new framework for deep representation learning based on spectral analysis with multiple layers.

\section{The Proposed SA-Net}
\label{sec:the-proposed-method}

\subsection{Main Idea}

At present, the dominating technique in deep representation learning  is CNN \cite{Wang2016LearningCA}, consisting of multiple layers, such as convolutional layers,  pooling layers, and softmax layers.  In a CNN, the convolution operation is the key to produce the representative and discriminative representations. 
Researchers also attempt to extend other techniques,  such as $ k $-means \cite{Coates2012Learning-deep-kmeans},  to a network structure for deep representation learning. 
In this paper, we propose a new representation learning method via expansion of the concept in spectral analysis, and to the best of our knowledge, we are the first to build a deep learning network by stacking spectral analysis procedures both consecutively and parallelly.

\begin{figure*}[htb]
	\centering
	\includegraphics[width=0.9\linewidth]{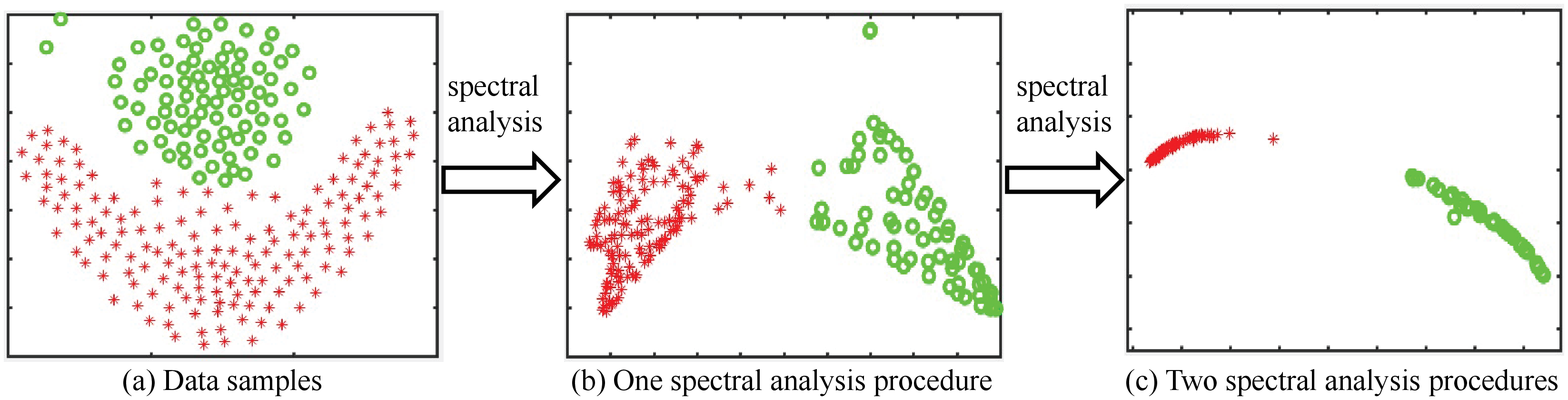}
	\caption{Illustration of samples in a two-cluster dataset and their corresponding spectral features, where  (a) the data samples; (b) the shallow spectral features of the samples obtained from one spectral analysis procedure;  and (c) the deep spectral features of the samples obtained from two consecutive spectral analysis procedures. }
	\label{fig:multipleSpectral}
\end{figure*}

It is widely recognized that, as relaxation of the cluster assignment vectors $ \hat{Y}=\{\hat{y_i}\}_{i=1}^N $,  the spectral features $ Y=\{y_i\}_{i=1}^N $ in algorithm \ref{alg:spectral-clustering} are more suitable for clustering than the original data points $ \{I_i\}_{i=1}^N $ \cite{vonLuxburg2007-a-tutorial-on-spectral-clustering}. In other words, a spectral analysis procedure can enhance the intra-cluster similarity as well as the inter-cluster separability. 
Inspired by such an observation, we consider to conduct multiple spectral analysis procedures \textbf{consecutively} to learn deep representations and hence create more spaces for research towards improved clustering performances.
An intuitive example is shown in Fig. \ref{fig:multipleSpectral}, where  part-(a) shows a two-cluster data set and part-(b)  shows the spectral features  obtained from a single spectral analysis procedure. By taking the features in part-(b) as the input, the second spectral analysis procedure produces a set of more clustering-friendly features (as shown in  part-(c)). Specifically, the deep features in part-(c) associate with a higher Calinski-Harabasz(CH) score than the shallow features in part-(b).

\begin{figure}[h]
	\centering
	\includegraphics[width=0.7\linewidth]{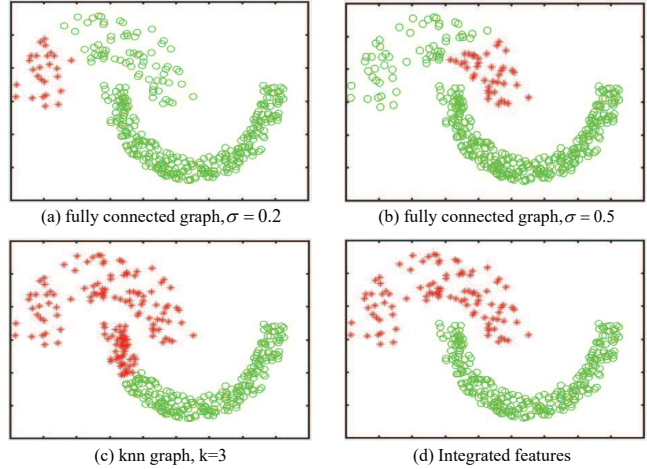}
	\caption{Illustration of a two-cluster dataset and the clustering results by different spectral analysis procedures, where a), b), and c) use a single type of spectral features, d) concatenates all of the three spectral features  before conducting $ k $-means. }
	\label{fig:combined_features}
\end{figure}

As previously mentioned, there are three different methods to construct the  affinity matrix, involving two different types of Laplacian matrices. Since each of them has its own advantage, and it remains difficult to determine which one to use for a given dataset, yet the variation of  a parameter can influence the clustering results significantly.
Fig. \ref{fig:combined_features} shows an example on two-cluster dataset and visualizes the clustering results of different spectral analysis procedures with a symmetrically normalized matrix.
In Fig. \ref{fig:combined_features}, part-(a) and part-(b) adopt a fully connected graph (i.e. $ w_{ij}=exp(-\|x_i-x_j\|^2/(2\sigma^2)) $) to construct the affinity matrix with  $ \sigma =0.2 $ and $ \sigma=0.5 $, respectively, and  part-(c) adopts knn to construct the affinity matrix with the parameter $ k =3 $. 
As seen, all these three spectral analysis procedures do not have optimal parameters and consequently,  each of them mis-cluster a portion of data points. 
In order to improve the clustering performance, we propose to conduct multiple spectral analysis procedures \textbf{parallelly} and integrate them together by simply concatenating their spectral features.
After applying a $ k $-means clustering on the concatenated features, we present our clustering results in part-(d). 
As seen, the integration of three spectral analysis procedures can cluster all the data points correctly, though none of them can achieve this individually. 
This indicates that we can improve the clustering performance by integrating  multiple spectral analysis procedures with sub-optimal affinity matrices. Similarly, different Laplacian matrices can also collaborate with each other to boost the clustering performance.
In this way, multiple parallel spectral analysis procedures can be elegantly integrated to work together collectively and collaboratively, and thus deep correlations across the data samples can be effectively identified and exploited for improved clustering performances.

\begin{figure*}[htb]
	\centering
	\includegraphics[width=0.99\linewidth]{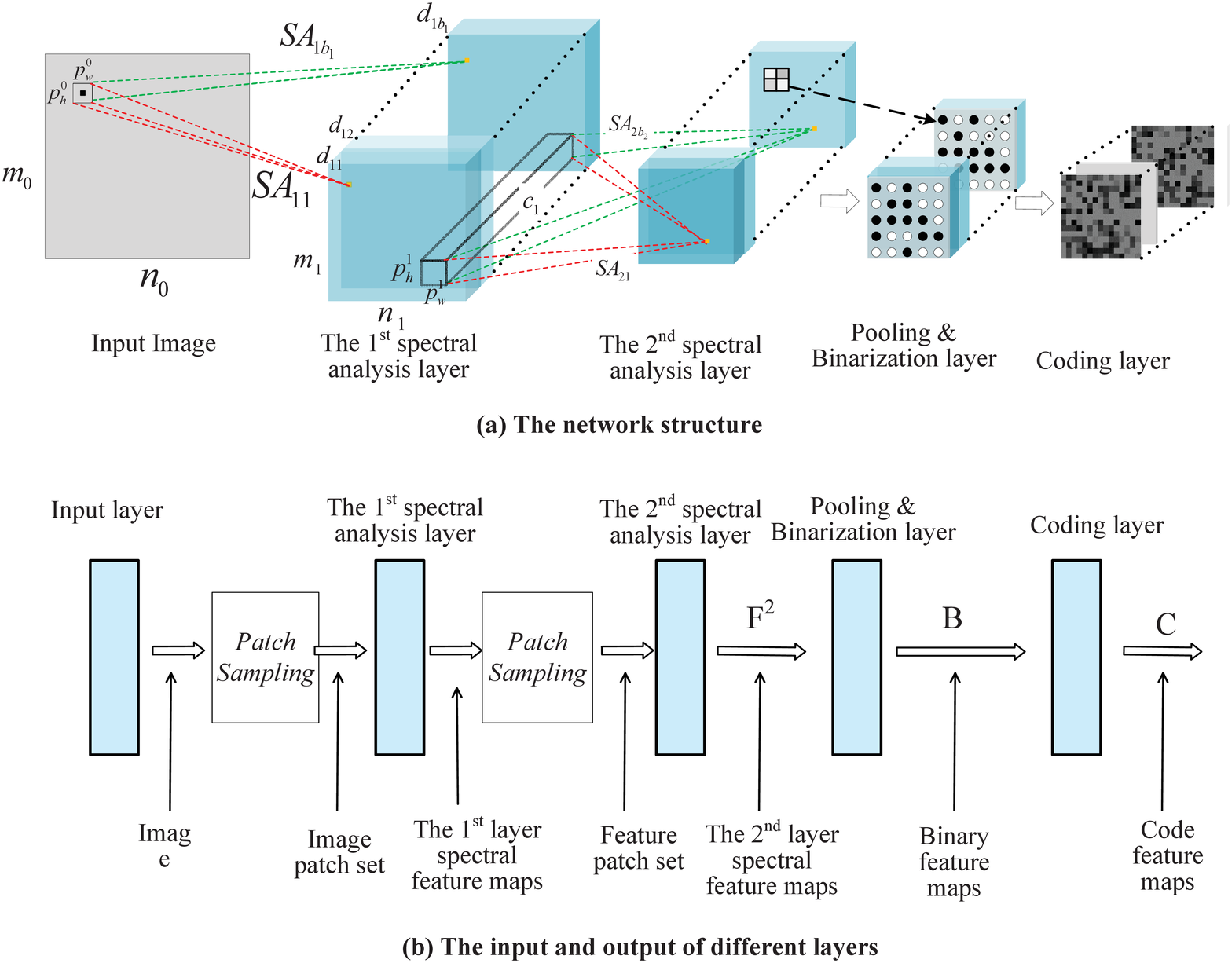}
	\caption{Illustration of the structure of the proposed SA-Net. In addition to the input and output layers, there are five layers in the proposed network, including two spectral analysis layers, one pooling layer, one binarization layer, and one coding layer. (a) shows the details of the whole network structure and (b) shows the input and output of different layers}
	\label{fig:network}
\end{figure*}

\subsection{SA-Net Structure}
\label{subsec-network-structure}

In this subsection, we propose a new network structure that can extract deep features or representations for the task of image clustering. Fig. \ref{fig:network} provides an overview of the proposed network structure. As seen, the proposed network consists of five layers, i.e. two spectral analysis layers, one pooling layer, one binarization layer, and one coding layer. Note that we can easily build  deeper networks by adding more layers out of  the four different types.

In order to enhance the clustering friendliness of its input, a spectral analysis layer produces the concatenation of spectral features obtained from multiple parallel spectral analysis procedures.  
For the convenience of description, the output features of a spectral analysis layer is referred to as spectral features.
In Algorithm \ref{alg:spectral-clustering}, $ y_i $ denotes the spectral feature of sample $ I_i $.
In general, spectral features are theoretically  more intra-cluster compact and inter-cluster separate \cite{vonLuxburg2007-a-tutorial-on-spectral-clustering}.
In practice, however,  different spectral analysis procedures have their own capabilities in dealing with data samples with various distributions.
In order to deal with data samples that follow different distributions, we extract a range of spectral features from multiple spectral analysis procedures and stack them together.
The first spectral analysis layer takes the image patches as the input and then its extracted spectral features are fed into the second layer for further processing to increase the discriminative power of the first layer spectral features.

The pooling layer summarizes the neighboring spectral features, and  takes the strongest response  to represent the visual appearance. The binarization layer binarizes the spectral features, and the coding layer transforms the binary feature maps into numerical feature maps, which provides better suitability and more friendliness for $ k $-means clustering.

Detailed descriptions and discussions of all these layers are provided as follows. 
For an image clustering task,  we assume that the image dataset  $I= \{ I_i | 1\leq i \leq N \}$ consists of $ N $  samples, and the image size is $ m_0\times n_0 $ with $ c_0 $ channels, i.e. $ I_i \in R^{m_0\times n_0\times c_0} $. 

(1) The first spectral analysis layer

The first layer conducts spectral analysis based on the densely sampled image patches.  
In the patch sampling procedure, we pad the image to include the border information. 
Around each of a subset (or all) pixels, we crop an image patch of size $ p_h^0 \times p_w^0 \times c_0 $.
With a stride of $ s_0 $ in patch sampling, we obtain $n_p^0= m_1 \times n_1= \lceil \left. m_0 \middle/ s_0 \right. \rceil \times \lceil \left. n_0 \middle/ s_0 \right. \rceil
$ patches in total from the image.   
In order to address the problem of illumination variation, a normalization procedure subtracts the mean from each image patch. 
For an image $ I_i $, we obtain a set of normalized image patches $ X_i^0=\{x_{ij}^0|1\leq j \leq n_p^0\} $. 
The patch set is $ X^0=\{X_1^0,X_2^0,\cdots, X_N^0\} $ with size of  $ N\times n_p^0 $.

Given the image patch set $ X^0 $, the $ t $th $  (1 \leq t \leq b_1) $ spectral feature in the $ 1 $st layer, i.e.  $ SA_{1t} $, extracts the spectral feature $ f_{ij}^{1t} \in R^{1 \times d_{1t}}$ for the  patch $ x_{ij}^0 $.
The parameter $  b_1 $ counts different spectral analysis procedures in the first layer and $ d_{1t} $ denotes the dimension of the spectral feature. 
Taking the normalized Laplacian matrix $ L_{sym} $ in Eq. (\ref{eq:symmeteric-Laplacian}) as an example, we can formulate the $ t $th spectral analysis procedure by the following objective function:
\begin{equation}
\label{eq-first-layer}
\max_{F_{1t}^T{D_t^1}{F_{1t}}=E_{d_{1t}}} tr(F_{1t}^T{W_t^1}F_{1t})
\end{equation} 
where $ {W_t^1} $ is the affinity matrix, $ {D_t^1} $ is a diagonal matrix with $ (D_t^1)_{ii}=\sum_{j=1}^{N} (W_t^1)_{ij} $, and $ F_{1t}= \begin{bmatrix}
{(f_{11}^{1t})}^T \quad {(f_{12}^{1t})}^T \cdots {(f_{Nn_p^0}^{1t})}^T
\end{bmatrix}^T  \in R^{(N\ast n_p^0) \times d_{1t}} $.
The following integrates the $ b_1 $ spectral analysis procedures into a single objective function:
\begin{equation}
\label{eq-first-layer-integration-objectie-function}
\max_{\substack{F_{1t}^T{D_t^1}{F_{1t}}=E_{d_{1t}} \\ t=1,2,...,b_1}}
tr
\left\{\begin{bmatrix}
F_{11}\\ 
F_{12}
\\ 
\vdots 
\\ 
F_{1b_{1}}
\end{bmatrix}^T
\begin{bmatrix}
W_1^1 & 0 & 0 &0 \\ 
0 & W^1_2 & 0 & 0\\ 
0 & 0 &  \ddots & 0\\ 
0&  0& 0 & W^1_{b_1}
\end{bmatrix}
\begin{bmatrix}
F_{11}\\ 
F_{12}
\\ 
\vdots 
\\ 
F_{1b_{1}}
\end{bmatrix}
\right\}
\end{equation} 
After obtaining $ F_{1t} \in R^{(N\ast n_p^0) \times d_{1t}} $, we stack the spectral features belonging to the same image into $ d_{1t} $ feature maps with the size of  $ m_1 \times n_1  $, instead of clustering them directly.
Note that, any one of the spectral analysis procedures in Eq. (\ref{eq-first-layer-integration-objectie-function}) can also be formulated by the left normalized Laplacian matrix $ L_{rw} $. In general, the $ b_1 $ different spectral analysis procedures can be different from each other in one or more of the following three aspects, i.e. the affinity matrix, the Laplacian matrix, and the computing method to produce the spectral features (further details are given in Sec. \ref{subsec-different-spectral-analysis-procedures}). 
With $ b_1 $ different  spectral analysis procedures, we obtain the spectral features of an image with the dimensionality of $ m_1 \times n_1 \times c_1 $, where $ c_1=\sum_{t=1}^{b_1}d_{1t} $ sums the dimensionality of $ b_1 $ different sets of spectral features.
Let $ F_i^1 \in R^{m_1 \times n_1 \times c_1} $ denote the first layer spectral features of the $ i $th image, and $ F^1=\{F_i^1| 1 \leq i \leq N \} $.

(2) The second spectral analysis layer 

This layer has two operational steps. The first step is to sample the  feature patches on the output of the first layer, i.e. $ F^1 $. Let the feature patch set be $ X^1 $, the second step is to conduct spectral analysis procedures on $ X^1 $ and produce the second layer spectral features $ F^2 $.

Let the size of the feature patches be $ p_h^1 \times p_w^1 \times c_1 $ as shown in  Fig. \ref{fig:network}, each feature patch carries the information learned from a larger patch with the size of $ (p_h^1+p_h^0-1) \times (p_w^1+p_w^0-1) \times c_0 $ in the original image. 
This indicates that, while the first layer deals with the correlations between small image patches, the second layer discovers the correlations among larger image areas from the original image.  
In addition, a feature patch also integrates the discriminative information learned by different spectral analysis procedures, which are suitable for the clustering of data samples with various distributions.
This layer provides an elegant manner to integrate different  spectral analysis procedures together in the feature-level, in order to enable them to work collaboratively.

(3) The pooling layer

The pooling layer summarizes the neighboring spectral features within the same spectral map, which can be conducted on the spectral features of the first or the second spectral analysis layer.  To illustrate the specific operation process, we take  spectral features of the second layer  as an example. 
The spectral analysis $ SC_{2t} $ produces $ d_{2t} $ different feature maps with the  size $ m_2 \times n_2 $ for each image, and each feature map is associated with an eigenvalue.
The pooling operation is conducted inside each feature map. 
For a $ s_p \times s_p $ grid centered at a point, the pooling operation only keeps the strongest response in terms of an absolute value, which can be  mathematically expressed as:
\begin{equation}
pooling(G)=g_{kl} \qquad where \quad |g_{kl}|=\max \limits_{ij} |g_{ij}|
\end{equation}
where $ g_{ij} $ denotes the feature in the $ i $th row and the $ j $th column of the target feature grid. The pooling grids can be overlapped in our algorithm.

(4) The binarization layer

From the graph cut point of view, the sign of spectral features (positive or negative) carries the cluster information \cite{vonLuxburg2007-a-tutorial-on-spectral-clustering}. 
For a clustering task with only two clusters, specifically, we can simply take a single eigenvector in step $ 3 $ of Algorithm \ref{alg:spectral-clustering}, and cluster  the data $ I_i $ into the first cluster if $ y_i >0 $ and   the second cluster if $ y_i<0 $.
This observation explicitly shows the importance of the sign of the spectral features in the clustering process. Following the pooling layer, correspondingly, we use a binarization layer to binarize the spectral features, in which $ B_{ij} $ denotes the $ j $th binary feature maps associating with the $ i $th image, 
and  $ B=\{B_{ij}\} $ denotes the binary feature map set.

\begin{figure}[htb]
	\centering
	\includegraphics[width=0.6\linewidth]{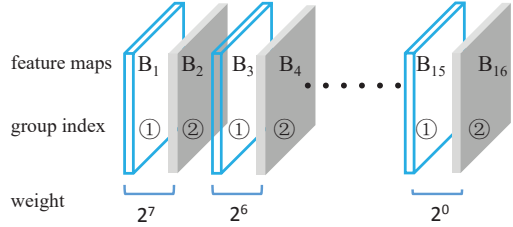}
	\caption{Illustration of partitioning the feature maps into groups, where the $ 16 $ feature maps are permuted based on their corresponding eigenvalues in a non-descending order, and we assign larger weights to those the feature maps with smaller eigenvalues. }
	\label{fig:coding}
\end{figure}

(5) The coding layer

Following the binarization layer, a coding layer \cite{chan-2015-tip-pcanet} is added to transform the binary code into decimal numbers and thus make it feasible for the following clustering operation. In this layer, we first partition the binary features of each image into different groups. Each group  consists of $ L $ binary feature maps, and  $ L $ is normally set to be $ 8 $.
With $ n_b $ binary feature maps, the coding layer produces  $ \lceil \left. n_b \middle/ L \right. \rceil
$ decimal feature maps. 
Let $ B_{ij}^k$ be the $ j $th $ (1\leq j \leq L) $ binary feature map  in the $ k $th group for the $ i $th image. 
At the position $ (u,v) $, we take the $ L $ binary features $ B_{ij}^k(u,v) $ and convert them into decimal by:
\begin{equation}
C_i^k(u,v)=\sum \limits_{j=1}^L 2^{L-j}B_{ij}^k(u,v)
\end{equation}
In this way, we obtain the $ k $th decimal feature map $ C_i^k $ for the $ i $th image. 
Note, we assign large weights to the spectral features with small eigenvalues, due to their strong discriminative capabilities.
By setting $ L $ to be $ 8 $,  we can produce the gray maps in the coding layer   as shown in Fig. \ref{fig:network}. 
The final clustering results are obtained by applying a simple $ k $-means to the output of the coding layer.
Fig. \ref{fig:coding} shows an example of partitioning $ 16 $ feature maps into two groups and assigning a weight to each feature map. In this figure, the feature maps are permuted based on their corresponding eigenvalues in a non-descending order, i.e.  $ v_i \leq v_{i+1}$, where $ v_i $ is the eigenvalue associated with $ B_i $.

\subsection{Operational set-up across different spectral analysis procedures}
\label{subsec-different-spectral-analysis-procedures}
In our proposed SA-Net, the  spectral analysis procedures are different from each other in terms of  affinity matrix, Laplacian matrix, and computing method. 
Tab. \ref{tab:different-spectral-analysis}  presents an overview of all these methods, from which it can be seen that we have three different affinity matrices and three types of Laplacian 
matrices to design the spectral analysis procedures. In addition, we have four different methods to compute the spectral features based on a given Laplacian matrix.  The last column of Tab. \ref{tab:different-spectral-analysis} provide the details of  whether we adopt the corresponding item or how we set the parameters.

\begin{table}[htb]
	\renewcommand{\arraystretch}{1.4} 
	\caption{Overview of spectral analysis procedures.}
	\begin{center}
		\begin{threeparttable}
			\begin{tabular}{c l l}
				\hline
				Stage & Methods & Settings\\
				\hline
				\multirow{3}{*}{Affinity  Matrix}   & $ k $-nearest-neighborhood & $ k=5,  9, 17, 21 $    \\ 
				& $ \epsilon $-nearest-neighborhood & $ \epsilon =0.5\eta,\eta,2\eta$\tnote{*}               \\ 
				& Fully connected matrix  &  $\sigma=10^{-1}, 10^{-2}, 10^{-3}$, or self-tunning  
		  \\ \hline
			\multirow{3}{*}{Laplacian Matrix} & L=D-W        & No\tnote{**}               \\ 
				& $ L_{sym}=E-D^{-1/2}WD^{-1/2}      $            & Yes           \\ 
				& $ L_{rw}=E-D^{-1}W $             & Yes              \\ \hline
				\multirow{4}{*}{Computing Method} & Eigen-decomposition      & No    \\
				& Lanczos method \cite{Saad2011Numerical} & Yes\\
				& Nystrom approximation \cite{Fowlkes-tpami2004-spectral-grouping} & Yes \\
				& Mini-batch analysis \cite{Han-ijcnn2017-miniBatch} & Yes \\ \hline
			\end{tabular}
			\begin{tablenotes}
				\item[*] $ \eta $ denotes the longest edge in the minimal spanning tree.
				\item[**] \textit{Yes} or \textit{No} denotes whether the corresponding term is used in this paper.
			\end{tablenotes}
		\end{threeparttable}
	\end{center}
	\label{tab:different-spectral-analysis}
\end{table}

To construct an affinity matrix  with $ k $-nearest-neighborhood, we choose a parameter $ k $ so that the affinity graph is connected. 
To determine the parameter $ \epsilon $ for $ \epsilon $-nearest-neighborhood affinity matrix construction, we first obtain a minimal spanning tree from the fully connected graph, then set the parameter $ \epsilon $ to be $ 0.5\eta, \eta$, or $ 2\eta $, where $ \eta $ denotes the longest edge in the minimal spanning tree.  
For a fully connected graph, we use empirical parameters 
or  the self-tuning method \cite{Zelnik-Manor:2004:Self-tuning} to determine an  appropriate parameter value for each sample point.

There are three different types of Laplacian matrices, including  one unnormalized and two normalized matrices.
We adopt the two types of normalized Laplacian matrices, i.e. $ L_{sym} $ and $ L_{rw} $, for our proposed SA-Net, as they have already shown their advantages in clustering. Theoretically, these two normalized Laplacian matrices implement the essential objective in maximizing both the inter-cluster separability and the intra-cluster similarity. As the unnormalized matrix (i.e. $ L $) only takes the first half of the objective into consideration, it is removed from our consideration.

Due to its high computational complexity, we do not apply the traditional eigen-decomposition procedure to our patch-based spectral analysis procedures.  As the Lanczos method \cite{Saad2011Numerical} has shown its advantages in decomposition of sparse matrices, we adopt this method to produce the spectral features from  the sparse affinity matrices constructed by $ k $-nearest-neighborhood and $ \epsilon $-nearest-neighborhood. 
The computational complexity of Lanczos method is $ O(N_{matrix} \cdot n_{eig}\cdot n_{iter}) $, where $ N_{matrix} $ denotes the width or height  of the Laplacian matrix, $n_{eig}$ denotes the number of eigenvectors to produce, and $n_{iter} $ denotes the number of iterations. 
For image clustering reported in this paper, the parameter $ N_{matrix} $ is a multiple of the number of images $ N $. For example, we have $ N_{matrix}=N\times n_p^{0} $ in the first spectral analysis layer, where $  n_p^{0}$ denotes the number of patches in each image.
As given in each experiment of sec. \ref{sec:experiment}, the parameter $n_{eig} $ varies in different spectral analysis procedures and is no larger than $ 64 $.  We set the parameter $ n_{iter}  $ to be $ 1000 $ in this paper. 
For the dense affinity matrix, we apply the Nystrom approximation-based method \cite{Fowlkes-tpami2004-spectral-grouping} or mini-batch spectral clustering (MBSC) \cite{Han-ijcnn2017-miniBatch} to produce the spectral features. 
The computational complexity of the Nystrom method is $ O(n_{eig} {\ell_{col}^2}+n_{eig} {\ell_{col}}N_{matrix}) $, where $ \ell_{col} $ denotes the number of representative columns sampled from the Laplacian matrix. In the implementation of Nystrom method, we set $ \ell_{col}= log N_{matrix} $  and adopt sparse matrix greedy approximation
(SMGA) sampling method \cite{smga-Smola:2000:SGM:645529.657980} to select the columns in a greedy manner. 
The  number of eigenvectors $ n_{eig} $, which equals to the  target rank, is given in each experiment.  Taking the face image clustering as an example,  we set $ n_{eig}=64 $ in the first spectral layer and $ n_{eig}=16 $ in the second spectral layer.
In the MBSC, we set the size of mini-batch to be $  N_{matrix}^{\frac{1}{2}}$, leading to the computational complexity of    $ O( N_{matrix} n_{eig}^2+N_{matrix}^2 n_{eig}+n_{eig}^3) $.

\section{Experiments}
\label{sec:experiment}

To evaluate the proposed SA-Net, we carry out extensive experiments on a range of image clustering tasks, including  handwritten digit image clustering, face image clustering, natural image clustering, and fashion image clustering. We show the robustness of our method against parameter variation in digit image clustering. We also show the performance variations of our method when the number of spectral analysis procedures changes in face image clustering.
Four popular standard metrics are adopted for measuring the clustering performances, which include  accuracy (ACC),  nomarlized mutual information (NMI) ,  adjust rand index (ARI), and F1-score (FS).

To benchmark our proposed SA-Net, we compare our proposed with $ 11 $ existing clustering algorithms, which cover almost all the representative existing state of the arts in image clustering. These include:  k-Means (KM), normalized cuts (N-Cuts) \cite{Shi2000-pami-Normalized}, self-tuning spectral clustering (SC-ST) \cite{Zelnik-Manor:2004:Self-tuning}, large-scale spectral clusteirng (SC-LS) \cite{Chen2011Large}, 
agglomerative clustering via path integral (AC-PIC) \cite{Zhang2013Agglomerative}, spectral embedded clustering (SEC) \cite{Nie2011Spectral}, local discriminant models and global integration (LDMGI) \cite{Yang2010Image}, NMF with deep model (NMF-D) \cite{Trigeorgis2014A}, deep embedding clustering (DEC) \cite{Xie:icml2016:Unsupevisd-deep-learning},  joint supervised learning (JULE) \cite{Yang-cvpr2016-joint}, and Deep embeded regularized clustering (DEPICT) \cite{Dizaji-iccv2017-deepClustering}.

\subsection{Handwritten Digit Image Clustering}
For the convenience of validation, we conduct the experiments on two popular handwritten digit image datasets, i.e. 
MNIST \cite{mnist-lecun-1998-gradient} and USPS\footnote{https://cs.nyu.edu/roweis/data.html}. 
The USPS dataset is a handwritten digits dataset produced by the USPS postal service. There are $ 11,000 $ images in this dataset, each  belonging to $ 10 $ different classes (i.e. from $ 0 $ to $ 9 $), and the image size
is $ 16 \times 16 $. 
The MNIST dataset  is one of the most popular image
datasets, widely used for deep learning based research. In total, this dataset consists of $ 70,000 $ images, $ 60,000 $ for training  and $ 10,000 $ for testing. Each image in the dataset represents a handwritten digit, from $ 0 $ to $ 9 $. We use all  the data samples in our experiments, and the images are centered with the size  $ 28 \times 28 $.

We take MNIST dataset as the example to show the implementation details, and it is similar for the USPS dataset.
In the first layer, we sample $ 11 \times 11 $ image patches  with a stride of $ 5 $ both vertically and horizontally. With padding in the border, we sample $6 \times 6=36 $ image patches from an  $ 28 \times 28  $ image.

For $ k $-nearest-neighborhood affinity matrix construction, we set the parameter $ k $ to be $ 9 $ and $ 17 $. As mentioned previously, we have three different settings for the value of $ \epsilon $ in $ \epsilon$-nearest-neighborhood affinity matrix construction, i.e. $ 0.5\eta $, $ \eta $, and $ 2\eta $,  where $ \eta $ denotes the longest edge in the minimal spanning tree. 
We also construct three different dense affinity matrices. While one dense matrix is determined by the self-tunning method \cite{Zelnik-Manor:2004:Self-tuning}, the other two are constructed based on the Gaussian function $ w_{ij}=exp(-||x_i-x_j||^2/(2\sigma^2)) $ with the parameter $ \sigma $ equals to $ 0.1 $ and $ 0.01 $, respectively. Thus, we have $ 5 $ different sparse affinity matrices and $ 3 $ different dense affinity matrices.

A symmetric Laplacian matrix is computed from each of the affinity matrices to yield  spectral features. 
We apply Lanczos to obtain spectral features from sparse affinity matrices, and mini-batch anlaysis to derive spectral features from dense affinity matrices. For each of the Laplacian matrix, we calculate $ n_{eig}=64 $ eigenvectors, i.e. the  dimensionality of spectral features is $ 64 $. 
In summary, the spectral features produced by the first layer is of the size $ 6 \times 6 \times 512 $ for each image, with $  64 $ dimensional spectral features for every $ 8 $ different Laplacian matrices.

\begin{figure}[htb]
	\centering
	\includegraphics[width=0.6\linewidth]{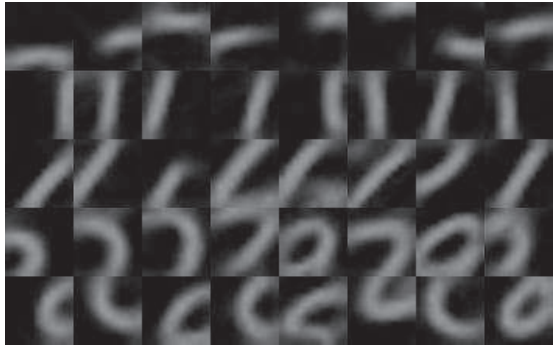}
	\caption{The typical visual patterns in the MNIST dataset}
	\label{fig:mnist-patches}
\end{figure}

To show that the first layer can learn the typical visual patterns in the image dataset, we conduct  a $ k $-means clustering based on the first layer spectral features and visualize $ 40 $ cluster centers in Fig. \ref{fig:mnist-patches}. As  seen, while the first three rows represent lines in different angles and positions, the last two rows represent different curve shapes  in the digit images.  
A proper combination of these visual patterns can produce a digit image.

In the second layer, we sample feature patches with the size of $ 4\times 4 \times 512 $.  With a stride of $ 1 $ and no feature padding, we obtain  $ 3\times3 $ feature patches for  
all the $ 6 \times 6 \times 512 $ spectral features at the first layer.
In other words, each image is associated with $ 9 $ feature patches with the dimensionality of  $ 4\times 4 \times 512 $.
Similarly, the second layer also uses both sparse affinity matrices and dense affinity matrices in the spectral analysis procedures.
By setting the parameter $ k $ to be $ 9 $ and $ 17 $, we use $ k $-nearest-neighborhood strategy to construct two sparse affinity matrices.
To maximize the strength of the multiple spectral analysis procedures, we also construct two dense affinity matrices by self-tuning method and the Gaussian distance with $ \sigma=0.1 $, respectively.
With a symmetric Laplacian matrix employed, therefore, this layer has $ 4 $ different spectral analysis procedures altogether.
As each spectral analysis produces $ n_{eig}=16 $ dimensional features, the second spectral analysis layer produces $ 64 $ feature maps for every image and each feature map is of the size $ 3\times3 $. 
Following the binarization and coding layer (with $ L=8 $), we now have $ 64/8=8 $ coding feature maps, each of which has the size of $ 3 \times 3 $. In other words, the dimensionality of the  features  for the final $ k $-means procedure is $ 72 $.

\begin{table}[]
	\caption{Comparative  results on handwritten digit image datasets}
	\begin{tabular}{lllll@{\hskip 0.3in}llll}
		\hline
		& \multicolumn{4}{c}{MNIST}    & \multicolumn{4}{c}{USPS}     \\ 
		\cline{2-9}
		& ACC   & ARI   & NMI   & FS    & ACC   & ARI   & NMI   & FS    \\
		\hline 
		KM     & 0.534 & 0.408 & 0.500 & 0.347 & 0.460 & 0.430 & 0.451 & 0.392 \\ 
		N-Cuts & 0.327 & 0.311 & 0.411 & 0.301 & 0.314 & 0.449 & 0.675 & 0.462 \\ 
		SC-ST  & 0.311 & 0.291 & 0.416 & 0.289 & 0.308 & 0.572 & 0.726 & 0.491 \\
		SC-LS  & 0.714 & 0.627 & 0.706 & 0.637 & 0.659 & 0.599 & 0.681 & 0.614 \\ 
		SEC    & 0.804 & 0.700 & 0.779 & 0.766 & 0.544 & 0.509 & 0.511 & 0.552 \\ 
		AC-PIC & 0.115 & 0.095 & 0.017 & 0.154 & 0.855 & 0.729 & 0.840 & 0.754 \\ 
		LDMGI  & 0.842 & 0.795 & 0.802 & 0.717 & 0.580 & 0.538 & 0.563 & 0.525 \\ 
		NMF-D  & 0.175 & 0.159 & 0.152 & 0.212 & 0.382 & 0.334 & 0.287 & 0.251 \\ 
		DEC    & 0.844 & 0.795 & 0.816 & 0.716 & 0.619 & 0.554 & 0.586 & 0.635 \\ 
		JULE   & 0.959 & 0.832 & 0.906 & 0.814 & 0.922 & 0.749 & 0.858 & 0.801 \\ 
		DEPICT & 0.965 & 0.808 & 0.917 & 0.851 & 0.964 & 0.846 & 0.927 & 0.809 \\ 
		SA-Net & 0.970 & 0.838 & 0.923 & 0.888 & 0.976 & 0.857 & 0.936 & 0.900 \\ \hline
	\end{tabular}
	\label{tab:handwritten-clustering-performance}
\end{table}

Tab. \ref{tab:handwritten-clustering-performance} lists the experimental results of our proposed SA-Net in comparison with the benchmarks. Across all  the four assessment metrics, 
our proposed SA-Net achieves the best performances indicating that the proposed network can learn feasible deep features for the clustering task.

In order to show the robustness of our method against parameter variations, we conduct  further experiments on the MNIST testing subset and list the results in table \ref{tab:mnist-testing-robustness}. In both of the two spectral analysis layers, we change the parameter $ k $ in k-nearest-neighborhood affinity matrix and parameter $ \sigma $ in the dense affinity matrix. 
The first layer has two k-nearest-neighborhood affinity matrices and two dense matrices determined by $ \sigma $. Thus, we have two values for $ k $ and two values for $ \sigma $ in the first layer. Similarly, we have two values for $ k $ and one value for $ \sigma $ in the second layer.  With three different settings for these parameters, the accuracy of our method only varies slightly, which are 0.974, 0.966, and 0.969, validating that our proposed SA-Net does achieve a good level of robustness against the parameter variations.

\begin{table}[htb] 
	\centering
	\caption{The accuracies achieved by SA-Net with different parameters. For simplicity,  the following abbreviations are adopted: SAL: spectral analysis layer; PL: pooling layer; BL: binarization layer; CL: coding layer. We also use \checkmark and  \texttimes \,  to indicate whether the layer is included or not in our experiments.}
	\begin{tabular}{p{3cm}p{3cm}p{0.5cm}p{0.5cm}p{0.5cm}p{0.5cm}}
		\hline 
		 The 1st SAL& The 2nd SAL & PL & BL & CL & ACC 
		\\ \hline
		  \multirow{4}{*}{$ k $=5,17;$ \sigma $=0.1,0.05}  & \multirow{4}{*}{$ k $=5,17;$ \sigma $=0.1} & \Checkmark     & \Checkmark & \Checkmark & 0.974 \\ 
		  &  & \Checkmark & \Checkmark & \XSolid & 0.957 \\
		  &  & \Checkmark & \XSolid & \XSolid & 0.891 \\
		  &  & \XSolid & \XSolid & \XSolid   & 0.884 \\
		 \hline
		 \multirow{4}{*}{$ k $=9,21;$ \sigma $=0.05,0.01}  & \multirow{4}{*}{$ k $=9,21;$ \sigma $=0.01} & \Checkmark     & \Checkmark & \Checkmark & 0.966 \\ 
		 &  & \Checkmark & \Checkmark & \XSolid & 0.942 \\
		 &  & \Checkmark & \XSolid & \XSolid & 0.903 \\
		 &  & \XSolid & \XSolid & \XSolid   & 0.876 \\
		 \hline
		 \multirow{4}{*}{$ k $=5,21;$ \sigma $=0.1,0.05}  & \multirow{4}{*}{$ k $=5,21;$ \sigma $=0.01} & \Checkmark     & \Checkmark & \Checkmark & 0.969\\ 
		&  & \Checkmark & \Checkmark & \XSolid & 0.934 \\
		 &  & \Checkmark & \XSolid & \XSolid & 0.912 \\
		 &  & \XSolid & \XSolid & \XSolid   & 0.906 \\
		 \hline
	 \end{tabular} 
 	\label{tab:mnist-testing-robustness}
 \end{table}

The experimental results in table \ref{tab:mnist-testing-robustness} also empirically show the effectiveness of the pooling layer, the binarization layer, and the coding layer.
When we remove the coding layer, as seen, the accuracy drops about $2\%  $ in all of the three settings. 
This is due to the fact that the coding layer assigns higher weights to the feature maps in order to increase the discriminating power, yet all of the feature maps have equal weights without the coding layer.
Although we may lose information in the operations of pooling and binarization, they can indeed improve the accuracy by reducing the intra-cluster distance and improving the cluster compactness.

\subsection{Face Image Clustering against occlusions}

To have more comprehensive evaluations upon our proposed SA-Net, we carry out another phase of experiments by testing our proposed against occlusions via  three publicly available  face image data sets AR \cite{ar-dataset}, YaleB \cite{Georghiades2002From-YaleB}, and CMU PIE \cite{cmu-pie-database}.  While the occlusions in YaleB-O and CMU PIE-O are introduced by ourselves, the occlusions in AR are real ones generated by scarf and glasses.

\begin{figure}[htb]
	\centering
	\includegraphics[width=0.6\linewidth]{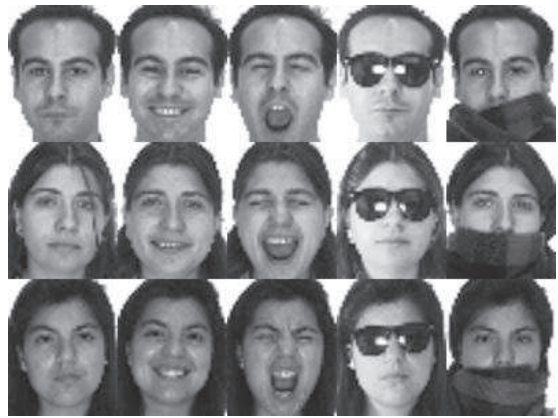}
	\caption{Illustration of example images from the AR dataset. }
	\label{fig:ar-images}
\end{figure}

The AR dataset \cite{ar-dataset} consists of more than $ 4,000 $ frontal face images from $ 126 $ people ($ 70 $ men and $ 56 $ women). The face images were captured under different conditions introduced by facial expression, illumination variation, and disguises (sunglasses and scarf). The images were captured in two sessions (with an interval of two weeks). In our experiment, we use a subset of the AR dataset, consisting of $  14 $ non-occluded images and $ 12 $ occluded images for each of the $ 120 $ persons. 
Fig. \ref{fig:ar-images} shows three groups of sample images out of $ 3 $ different persons.

The YaleB facial image dataset \cite{Georghiades2002From-YaleB} has around 64 near frontal images from $ 38 $ individuals. The images are captured under different illuminations.
Totally, this dataset has $ 2414 $ face images. 
The CMU PIE dataset \cite{cmu-pie-database} is collected by Carnegie Mellon University. The face images in this dataset are captured with different poses, illumination conditions and expressions. We use $ 2,856 $ images from $ 68 $ persons.

\begin{figure}[htb]
	\centering
	\includegraphics[width=0.5\linewidth]{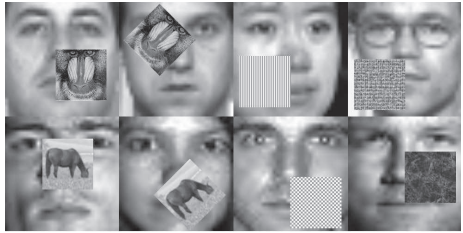}
	\caption{Illustration of eight example face images from YaleB and CMU PIE whose twenty percent pixels are occluded. The occlusions are not related to the face images and are randomly located. }
	\label{fig:occluded-yaleb-cmu-images}
\end{figure}

Compared with the AR dataset, neither YaleB nor CMU PIE has occluded face images. To test the robustness of our proposed SA-Net, we create two datasets with occlusions, i.e. CMU PIE occlusion (CMU PIE-O) and YaleB occlusion (YaleB-O).
In these two occluded datasets, we simulate contiguous occlusions by hiding $ 20\% $ of pixels at a randomly selected location with a block out of another irrelevant image, some samples of which are shown in Fig. \ref{fig:occluded-yaleb-cmu-images}.

\begin{table}[htb]
	\centering
	\caption{Comparative  results on AR face image dataset}
	\begin{tabular}{lllll|lllll}
		\hline
		& ACC   & ARI   & NMI   & FS    & & ACC   & ARI   & NMI   & FS    \\
		\hline 
		KM     & 0.534 & 0.408 & 0.500 & 0.347 & LDMGI &0.460 & 0.430 & 0.451 & 0.392 \\ 
		N-Cuts & 0.327 & 0.311 & 0.411 & 0.301 &  NMF-D&  0.314 & 0.449 & 0.675 & 0.462 \\ 
		SC-ST  & 0.311 & 0.291 & 0.416 & 0.289 & DEC    &0.308 & 0.572 & 0.726 & 0.491 \\
		SC-LS  & 0.714 & 0.627 & 0.706 & 0.637 & JULE   &0.659 & 0.599 & 0.681 & 0.614 \\ 
		SEC    & 0.804 & 0.700 & 0.779 & 0.766 & DEPICT & 0.544 & 0.509 & 0.511 & 0.552 \\ 
		AC-PIC & 0.115 & 0.095 & 0.017 & 0.154 & SA-Net &0.855 & 0.729 & 0.840 & 0.754 \\ 
		\hline
	\end{tabular}
	\label{tab:AR}
\end{table}

\begin{table}[htb]
	\caption{Comparative  results on CMU PIE and PIE-O face image datasets}
	\begin{tabular}{lllll@{\hskip 0.3in}llll}
		\hline
		& \multicolumn{4}{c}{CMU PIE}    & \multicolumn{4}{c}{CMU PIE-O}     \\ 
		\cline{2-9}
		& ACC   & ARI   & NMI   & FS    & ACC   & ARI   & NMI   & FS    \\
		\hline 
		KM     &    0.196 	&    0.190	&    0.266	&    0.153	&    0.158	&    0.154	&    0.258	&    0.123	\\
		N-Cuts &    0.129	&    0.156	&    0.223	&    0.111	&    0.105	&    0.128	&    0.211	&    0.081	\\
		SC-ST  &    0.218 	&    0.198	&    0.295	&    0.250	&    0.185 	&    0.170 	&    0.270	&    0.224 	\\
		SC-LS  &    0.282 	&    0.191 	&    0.277	&    0.265	&    0.246	&    0.169	&    0.189	&    0.242 	\\
		SEC    &    0.112	&    0.122	&    0.126	&    0.149	&    0.079	&    0.086	&    0.105	&    0.086	\\
		AC-PIC &    0.244	&    0.320	&    0.221	&    0.189	&    0.224	&    0.283	&    0.194	&    0.169	\\
		LDMGI  &    0.256	&    0.194	&    0.242	&    0.257	&    0.198	&    0.158	&    0.214	&    0.223	\\
		NMF-D  &    0.310	&    0.357	&    0.380	&    0.292	&    0.280	&    0.324	&    0.266	&    0.266 	\\
		DEC    &    0.421	&    0.389	&    0.477	&    0.348	&    0.403 	&    0.334	&    0.347	&    0.291	\\
		JULE   &    0.550	&    0.437	&    0.521 	&    0.497	&    0.522	&    0.421	&    0.311	&    0.423	\\
		DEPICT &    0.535	&    0.484	&    0.488 	&    0.453	&    0.516	&    0.374	&    0.320 	&     0.347	\\
		SA-Net &    0.610	&    0.497	&    0.605	&    0.537	&    0.569	&    0.423	&    0.567	&    0.518	\\ \hline
	\end{tabular}
	\label{tab:CMU-PIE}
\end{table}

\begin{table}[htb]
	\caption{Comparative  results on YaleB and YaleB-O face image datasets}
	\begin{tabular}{lllll@{\hskip 0.3in}llll}
		\hline
		& \multicolumn{4}{c}{YaleB}    & \multicolumn{4}{c}{YaleB-O}     \\ 
		\cline{2-9}
		& ACC   & ARI   & NMI   & FS    & ACC   & ARI   & NMI   & FS    \\
		\hline 
		KM     & 0.234  	& 0.208    &0.287   & 0.216 	& 0.117		& 0.070 	& 0.111		& 0.083 \\
		N-Cuts & 0.163   	& 0.139    &0.219  	&0.160 		&0.093 		& 0.113    	& 0.163 	&0.150\\
		SC-ST  & 0.308  	&0.276    & 0.268 	& 0.321 	&0.216  	&  0.170    &0.236  	& 0.197\\
		SC-LS  & 0.577    	&0.502    & 0.642  &0.547 		&0.167		&0.108    	&0.194   	& 0.125\\
		SEC    &  0.401   	&0.438    & 0.431  &0.452 		& 0.230  	&  0.158    & 0.157  	& 0.152\\
		AC-PIC & 0.472    	&0.482    &0.412 	&0.428   	&0.264  	& 0.208    	& 0.242  	&0.241\\
		LDMGI  & 0.574 		&0.597    &0.615	&0.509		& 0.230  	&0.176    	&0.172  	& 0.223 \\
		NMF-D  & 0.578  	&0.520    & 0.637	&0.613 		& 0.393  	&0.296    	& 0.310  	&0.242 \\
		DEC    & 0.571  	& 0.597   &0.569	&0.606		&0.402  	&  0.245   	&0.469  	&  0.441\\
		JULE   & 0.610 		&0.573    &0.697	& 0.649		& 0.471  	&  0.460    &0.496  	& 0.358\\
		DEPICT & 0.678  	& 0.619    & 0.650	&0.617		&0.418  	&0.393    	&0.441 		&0.380\\
		SA-Net & 0.766  	&0.648	  &0.707 	&0.681		&  0.621  	& 0.526 	& 0.580 	&0.497\\ \hline
	\end{tabular}
	\label{tab:YaleB}
\end{table}

Following the same design as for the clustering of digit images, we apply the proposed SA-Net shown in Fig. \ref{fig:network} to face image clustering. While each spectral analysis procedure produces  $ n_{eig}=64 $ eigenvectors in the first layer, the second layer produces $ n_{eig}=16 $ eigenvectors.
However, some of the implementation details need to be adapted correspondingly.  Firstly, for face image clustering, we sample patches of size $ 15 \times 15 $ with stride of $ 7 $. The patches are  larger than the ones used in the handwritten image dataset, in order to allow the typical visual patterns to cover meaningful parts of the faces. Secondly, we use the Nystrom approximation method (a different method from the digit image clustering experiment) to compute the spectral features from the dense affinity matrices. 
It is observed in our experiments that the computing method has little impact upon the clustering performances.

\begin{figure*}[h]
	\centering
	\includegraphics[width=0.8\linewidth]{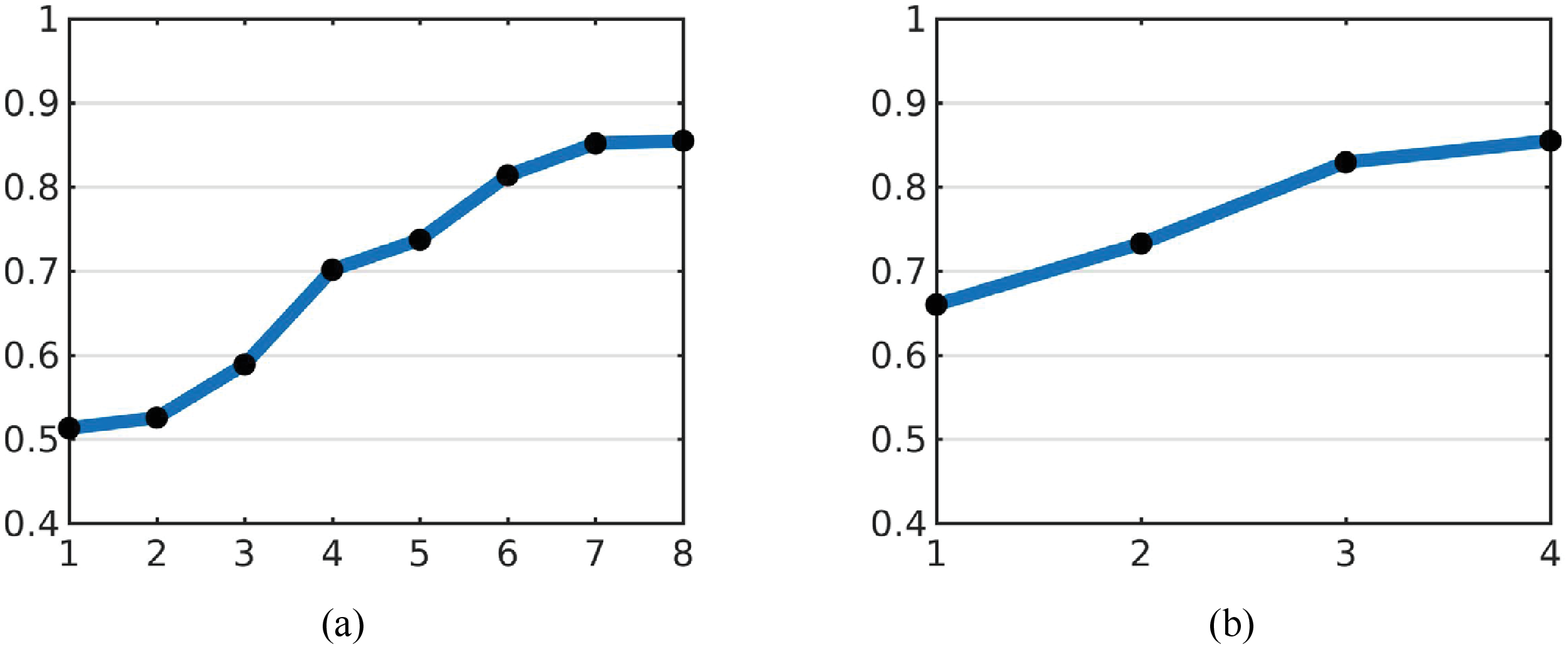}
	\caption{The average accuracy (vertical) versus the number of spectral analysis procedures (horizontal) in the first layer (a)  and  the second layer (b)}
	\label{fig:accuracy-versus-numberofSC}
\end{figure*}

Table \ref{tab:AR}, \ref{tab:CMU-PIE}, and \ref{tab:YaleB}  summarize all the experimental results, from which it can be seen that our proposed SA-Net outperforms all the $ 11 $ benchmarks selected out of the existing clustering algorithms. Further examinations of the experimental results also reveal that,  compared with all other spectral analysis-based clustering methods, the proposed SA-Net achieves additional advantages  in dealing with the occluded face images, due to the fact  that the network allows us to identify the local similarity between the face images at patch-level. In contrast, the existing spectral analysis-based methods only consider the global similarity between face images, yet the occlusion can significantly reduce the similarity between two images, even when they associate with the same person. This explains why the performances of the existing methods drop significantly on both Yale-O and CMU PIE-O.

In order to show that every spectral analysis procedure contributes positively to the clustering task, we adopt different number of spectral analysis procedures in our experiments and show the average accuracy in Fig. \ref{fig:accuracy-versus-numberofSC}.
Specifically, we keep all of the spectral analysis procedures in one layer and remove one or more spectral analysis procedures in the other layer. In Fig. \ref{fig:accuracy-versus-numberofSC} (a), we use 1 to 8 spectral analysis procedures in the first layer and 4 spectral analysis procedures in the second layer. In Fig. \ref{fig:accuracy-versus-numberofSC} (b), we use  8 spectral analysis procedures in the first layer and 1 to 4 in the second layer.
As  seen, the average accuracy increases when we use more spectral analysis procedures. 
This means that, as more spectral analysis procedures are added with our proposed SA-Net, more discriminative spectral features and subspaces are brought in to increase the discriminating power of our proposed method. 
In other words, the additional spectral analysis procedures  bring beneficial  subspaces for clustering, indicating that the parallel spectral analysis procedures are indeed working together collaboratively and collectively.
In principle, we can adopt more spectral analysis procedures, but the increase on clustering accuracies remains trivial, revealing that our choice of  $ 8 $ spectral analysis procedures in the first layer and $ 4 $ in the second layer  is sufficient not only in discovering clustering-friendly representations, but also in representing an appropriate balance between the computing cost and the effectiveness.

\subsection{Natural Image Clustering}
\label{sec:natural-image-clustering}

To assess how our proposed SA-Net performs on natural image clustering, we  further conduct  experiments on three more image  datasets,  STL-10, CIFAR-10, and CIFAR-100. They respectively have 13k, 60k, and 60k images, and the number of clusters for them are respectively $ 10 $, $ 10 $, and $ 20 $. 
For training purposes, we resize the images from STL-10 to $ 32 \times 32 \times 3 $, and let them be the same size as the images from the other two datasets.

\begin{table}[htb]
	\centering
	\caption{Results of SA-Net and baselines on STL-10, CIFAR-10, and CIFAR-100}
	\label{tab-natural-image-datasets}
	\small
	\begin{tabular}{llccccccc}
		\hline
		&     & SEC   & DEC   & JULE  & {\footnotesize DEPICT} & SA-Net-2& SA-Net-3& SA-Net-5 \\
		\hline
		\hline
		\multirow{4}{*}{{ STL-10}}    & ACC & 0.148 & 0.276 & 0.182 & 0.195  & 0.317 & \textbf{0.331} &0.328
		\\
		& ARI &  0.125 & 0.184  &  0.207 & 0.231  & 0.274 & \textbf{0.304} & 0.295
		\\
		& NMI & 0.212 & 0.359 & 0.277 & 0.264  & \textbf{0.398} & 0.351 & 0.386
		\\
		& {FS} & 0.216  & 0.276  &  0.237 & 0.245  &  0.327 & 0.330 & \textbf{0.353}
		\\
		\hline		
		\hline
		\multirow{4}{*}{{CIFAR-10}}  & ACC & 0.168  & 0.257 & 0.192 & 0.244  & 0.322 & 0.297 & \textbf{0.336}
		\\
		& ARI & 0.175  & 0.206  &  0.291 &  0.273 & 0.316 & \textbf{0.342} & 0.318
		\\
		& NMI & 0.204  & 0.301 & 0.272 & 0.295  & 0.341 & 0.356 & \textbf{0.374}
		\\
		& {FS} &   0.237  & 0.221& 0.268 & 0.217  & 0.283 & 0.316 & \textbf{0.323}\\
		\hline
		\hline
		\multirow{4}{*}{{CIFAR-100}} & ACC & 0.094  & 0.136 & 0.103 & 0.187  & 0.208 & 0.223 & \textbf{0.254}
		\\
		& ARI & 0.141  & 0.154  & 0.169  &  0.184 & 0.194 & 0.222 & \textbf{0.240}
		\\
		& NMI & 0.123  & 0.185 & 0.137 & 0.240  & 0.274 & 0.276 & \textbf{0.293}
		\\
		& { FS} & 0
		106  &  0.162 & 0.173  & 0.200  & 0.207 & 0.246 & \textbf{0.273}\\
		\hline
	\end{tabular}
\end{table}

We compare the proposed SA-Net with four representative deep learning methods, i.e. SEC \cite{Nie2011Spectral},  DEC \cite{Xie:icml2016:Unsupevisd-deep-learning},  JULE \cite{Yang-cvpr2016-joint}, and DEPICT \cite{Dizaji-iccv2017-deepClustering}.  	Based on the four types of layers (i.e. SAL-Spectral Analysis Layer, PL-Pooling Layer, BL-Binarization Layer, and CL-Coding Layer), we build three different network structures, including SA-Net-2 (\textit{SAL-SAL-PL-BL-CL}), SA-Net-3 (\textit{SAL-SAL-PL-SAL-PL-BL-CL}), and SA-Net-5 (\textit{SAL-SAL-PL-SAL-PL-SAL-PL-SAL-PL-BL-CL}).

In the first spectral layer, 
we sample $ 11 \times 11 \times 3 $ image patches with a stride of $ 2 $ pixels both vertically and horizontally. For each image of size $ 32 \times 32 \times 3 $, we obtain $ 16 \times 16 =256$ image patches. 
To deal with these difficult image datasets, we use $ 16 $ different spectral analysis procedures, involving $ 8 $ different affinity matrices and $ 2 $ normalized Laplacian matrices. The $ 8 $ different affinity matrices are: three k-nearest-neighborhood affinity matrices with $ k =9, 17, 21$; three $ \epsilon$-nearest-neighborhood affinity matrices with $ \epsilon=0.5\eta, \eta, 2\eta $; and two fully connected affinity matrices determined by a self-tunning method and  $ \sigma = 0.1$. Each spectral analysis procedure produces  spectral features with dimensionality of $ n_{eig}=16 $, and thus the dimensionality of the first layer spectral feature for an image is $ 16 \times 16 \times 256 $.

In the second spectral analysis layer, we sample $ 5\times 5 \times 256  $ feature patches and set the stride to be $ 2 $. With $ 8 $ different spectral analysis procedures each contributing $ n_{eig}=8 $ spectral features, we obtain the second layer spectral feature with a dimensionality of $ 8 \times 8 \times 64 $. In each of the following (i.e. the third, forth, and fifth) spectral analysis layers, we set the patch size to be $ 3\times 3 $ and adopt $ 8 $ different spectral analysis procedures. After pooling, binarization and coding, the dimensionality of the final feature map  for an image is $ 4 \times 4 \times 8 =128 $.

Tab. \ref{tab-natural-image-datasets} summarizes the experimental results for both the benchmarks and our  methods, where the best performances are highlighted in bold. As seen, the proposed methods outperform the benchmarks in terms of ACC, ARI, NMI, and FS. 
While SA-Net-5 achieves the best performance in $ 8 $ cases, SA-Net-3 and SA-Net-2 achieve the best in 3 cases and 1 case, respectively.
In the most difficult  CIFAR-100 dataset, the deepest SA-Net-5 achieves the best performances across all of the four evaluating metrics. 
Taking FS as the example, SA-Net-5 beats SA-Net-2 by a margin of $ 0.066 $ on CIFAR-100.
In terms of NMI, even the shallowest network SA-Net-2 can outperform  the existing benchmarks by a margin larger than $ 0.03 $ on all of the three datasets.  In terms of ACC (or FS), both SA-Net-3 and SA-Net-5 outperform the existing benchmarks by a margin larger than  $ 0.035 $ (or $ 0.045 $) on all of the three datasets.

\begin{figure*}[htb]
	\centering
	\includegraphics[width=0.8\linewidth]{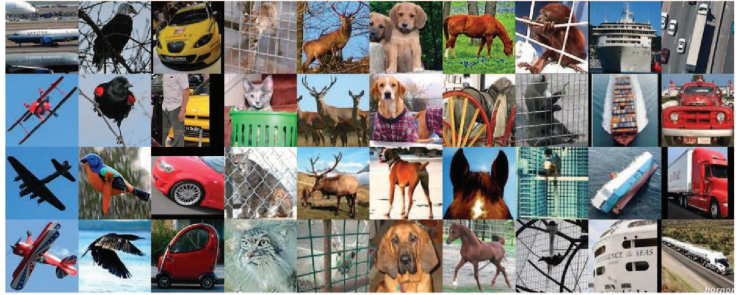}
	\caption{The STL-10 images which are far away from their associating cluster centers }
	\label{fig:stl-images}
\end{figure*}

To show the difficulty of these clustering tasks, Fig. \ref{fig:stl-images} illustrates some sample images from STL-10. As seen, the samples are indeed far away from  their associating cluster centers. While the \textit{planes} are captured under different views in the first column, the \textit{cats} are heavily occluded in the fourth column and the \textit{horses} are captured under quite different backgrounds in the eighth column.

\subsection{Fashion image clustering}

To test our proposed SA-Net for its capability in clustering variety of images with different styles, we carry out one more phase of experiments to
cluster the fashion images into different styles on the dataset HipsterWars \cite{hipsterwars-kiapour-eccv2014}. 
This dataset consists of $ 1,893 $ fashion images, each associating with one of five style categories, including hipster, bohemian, pinup, preppy, and goth. The numbers of images in these five categories are $ 376 $, $ 462 $, $ 191 $, $ 437 $, and $ 427 $ respectively. 
For the convenience of implementation without losing generality, we resize all the images into $ 600\times 400 \times3 $.

As in Sec. \ref{sec:natural-image-clustering}, we also adopt three different network structures, i.e. SA-Net-2, SA-Net-3, and SA-Net-5.
We compare our method with three existing state of the art benchmarks, including StyleNet \cite{Simo-fashion-cvpr2016}, ResNet \cite{He2016Deep}, and PolyLDA \cite{Hsiao-iccv2017}. 
The StyleNet is a  network for clothing that is trained from the Fashion 144K dataset \cite{SimoSerraCVPR2015}, and the ResNet is a popular network for image classification. We extract features from these two networks and obtain the clusters by $ k $-means. 
PolyLDA (polylingual Latent Dirichlet Allocation) is a  Bayesian nonparametric model to characterize the styles by discovering the compositions of lower-level visual cues.

In this experiment, the first spectral analysis layer samples image patches of size $ 32 \times 32 $. 
The main goal of the first layer is to  discover the typical visual patterns that appears in many fashion images, where we use the $ 8 $ different spectral analysis procedures as in the digit image clustering to learn the spectral features. 
In the second and the subsequent spectral analysis layers, we only use sparse affinity matrix constructed by the $ k $-nearest-neighborhood, with the parameter $ k $ equals to $ 5, 9, 17 $ and $ 21 $, respectively. We adopt both the symmetric normalized matrix and the left normalized matrix, and apply the Lanczos method for Laplacian matrix decomposition to produce spectral features. 
A spectral analysis procedure produces $ n_{eig}=64 $  eigenvectors
in the first and  $ n_{eig}=16 $ eigenvectors in the second or the subsequent spectral analysis layers.

\begin{table}[h]
	\centering
	\caption{Comparative  results  on HipsterWars dataset}
	\label{tab:hipsterwards-nmi-acc}
	\begin{tabular}{lcccccc}
		\hline
		& StyleNet & ResNet & PolyLDA & SA-Net-2 & SA-Net-3 & SA-Net-5 \\
		\hline	\hline
		ACC & 0.39     & 0.30   & 0.50    & 0.54  & 0.54 & 0.55\\
		ARI &  0.14   & 0.12  &   0.18  & 0.22 & 0.24 & 0.25 \\
		NMI & 0.20     & 0.16   & 0.21    & 0.20 & 0.20 & 0.21\\
		FS &   0.30   &  0.28  &   0.33 & 0.39 & 0.41 & 0.41\\
		\hline
	\end{tabular}
\end{table}

\begin{table}[hbt]
	\centering
	\caption{The confusion matrix of the proposed SA-Net-2 on the Hipsterwars dataset}
	\label{tab:confusion-matrix-hipsterwars}
	\begin{tabular}{lccccc}
		\hline
		& Hipster & Bohemian & Pinup & Preppy & Goth \\
		\hline
		Hipster  & 140      &  83      & 21      & 81      & 51  \\
		Bohemian & 47      & 328      & 18      & 28      & 41   \\
		Pinup   & 30      & 25      & 72      & 23      & 41   \\
		Preppy  & 92      & 49      & 35     & 202      & 59 \\
		Goth  & 64      & 20      & 28      & 32      & 283
		\\
		\hline
	\end{tabular}
\end{table}

As seen from Tab. \ref{tab:hipsterwards-nmi-acc},  our  method performs  better than the existing benchmarks in the four evaluating metrics. 
Specifically, our proposed method can improve the ACC by a margin of $ 5\% $, the ARI by a margin of $ 7\% $, the FS by a margin of $ 8\% $.	
As seen, there exist no significant differences among   the  performances of  our three network structures, indicating that two spectral analysis layers are sufficient for this small dataset. 

\begin{figure}
	\centering
	\includegraphics[width=0.6\linewidth]{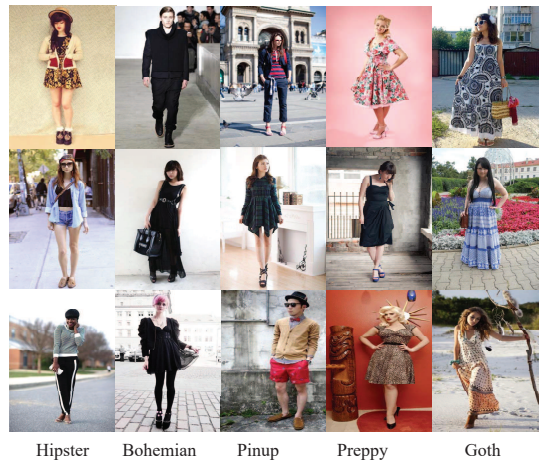}
	\caption{Illustration of the fashion image samples nearest to the cluster centers of the five different styles}
	\label{fig:representativeFashion}
\end{figure}

For the convenience of further examination and analysis, Tab. \ref{tab:confusion-matrix-hipsterwars} illustrates the values of our confusion matrix for SA-Net-2, and Fig. \ref{fig:representativeFashion} illustrates some image samples that are nearest to the cluster centers.

In addition, further examination reveals that all the compared methods fail to  achieve good  clustering results on this dataset. This is mainly due to two reasons.  Firstly, the images are obtained from on-line, not captured under any controlled environment, and as a result, they  are significantly different from each other in terms of the capturing view, illumination, and background (see Fig. \ref{fig:representativeFashion}).  Secondly, a style (and the resulting cluster) normally represents the coherent latent appearance between different parts of the fashion images, not simply a composition of several visual components. In other words,  the foregrounds of two images can  be quite different  even though they are from the same cluster.

\section{Conclusions}
\label{sec:conclusion}

In this paper, we have described a new deep learning network SA-Net for image clustering  based on the technique of spectral analysis. This provides one more method 
for deep representation learning, in addition to the popular convolutional neural network.
Our proposed network structure has four type of layers, including spectral analysis layer, binarization layer, coding layer, and pooling layer. Compared with the existing spectral analysis clustering methods, SA-Net achieves three advantages.
Firstly, while the existing spectral clustering methods learn representations by a single spectral analysis procedure, our proposed SA-Net conducts multiple  procedures in both consecutive and parallel manner to learn more clustering-friendly representations.  
Secondly, SA-Net can elegantly integrate different spectral analysis procedures and thus capable of dealing with different data sample sets.
Thirdly, by conducting spectral analysis procedures on image patches, SA-Net can discover the local similarity among images at patch-level, and hence it is more robust against occlusions than the existing spectral clustering methods.
Extensive experiments validate the effectiveness of the proposed SA-Net on a range of different image clustering tasks, including handwritten digit image clustering, face image clustering, natural image clustering, and fashion image clustering.

\section*{Acknowledgment}

The authors wish to acknowledge the financial support from: (i) Natural Science Foundation China (NSFC) under the Grant No. 61620106008; (ii) Natural Science Foundation China (NSFC) under the Grant No. 61802266; and (iii) Shenzhen Commission for Scientific Research \& Innovations under the Grant No. JCYJ20160226191842793.

\section*{References}

\bibliography{mybibfile}

\begin{thebibliography}{10}
\expandafter\ifx\csname url\endcsname\relax
  \def\url#1{\texttt{#1}}\fi
\expandafter\ifx\csname urlprefix\endcsname\relax\def\urlprefix{URL }\fi
\expandafter\ifx\csname href\endcsname\relax
  \def\href#1#2{#2} \def\path#1{#1}\fi

\bibitem{XU-survey-tnn2005}
R.~Xu, D.~Wunsch, Survey of clustering algorithms, IEEE Transactions on Neural
  Networks 16~(3) (2005) 645--678.

\bibitem{Yang2010Image}
Y.~Yang, D.~Xu, F.~Nie, S.~Yan, Y.~Zhuang, Image clustering using local
  discriminant models and global integration, TIP 19~(10) (2010) 2761--2773.

\bibitem{Trigeorgis2014A}
G.~Trigeorgis, K.~Bousmalis, S.~Zafeiriou, B.~W. Schuller, A deep semi-nmf
  model for learning hidden representations, in: ICML, 2014, pp. 1692--1700.

\bibitem{CHEN2018177-neuro}
H.~Chen, W.~Wang, X.~Feng, R.~He, Discriminative and coherent subspace
  clustering, Neurocomputing 284 (2018) 177 -- 186.

\bibitem{Wu-trans2015-fuzzy-anew}
C.~H. Wu, C.~S. Ouyang, L.~W. Chen, L.~W. Lu, A new fuzzy clustering validity
  index with a median factor for centroid-based clustering, IEEE Transactions
  on Fuzzy Systems 23~(3) (2015) 701--718.

\bibitem{Liu-2017pami-hierarchical}
A.~A. Liu, Y.~T. Su, W.~Z. Nie, M.~Kankanhalli, Hierarchical clustering
  multi-task learning for joint human action grouping and recognition, IEEE
  Transactions on Pattern Analysis and Machine Intelligence 39~(1) (2017)
  102--114.

\bibitem{Zhang2013Agglomerative}
W.~Zhang, D.~Zhao, X.~Wang, Agglomerative clustering via maximum incremental
  path integral, Pattern Recognition 46~(11) (2013) 3056--3065.

\bibitem{TONG20182355-neurocomputing}
Q.~Tong, X.~Li, B.~Yuan, Efficient distributed clustering using boundary
  information, Neurocomputing 275 (2018) 2355 -- 2366.

\bibitem{CAI2018316-neurocomputing}
Y.~Cai, Y.~Jiao, W.~Zhuge, H.~Tao, C.~Hou, Partial multi-view spectral
  clustering, Neurocomputing 311 (2018) 316 -- 324.

\bibitem{LEE2018210-neuro}
Y.~Lee, J.~Im, S.~Cho, J.~Choi, Applying convolution filter to matrix of
  word-clustering based document representation, Neurocomputing 315 (2018) 210
  -- 220.

\bibitem{ZHAO2018227-neurocomputing}
Y.~Zhao, Y.~Yuan, F.~Nie, Q.~Wang, Spectral clustering based on iterative
  optimization for large-scale and high-dimensional data, Neurocomputing 318
  (2018) 227 -- 235.

\bibitem{Chen2011Large}
X.~Chen, D.~Cai, Large scale spectral clustering with landmark-based
  representation, in: AAAI, 2011, pp. 313--318.

\bibitem{Ng01onspectral}
A.~Y. Ng, M.~I. Jordan, Y.~Weiss, On spectral clustering: Analysis and an
  algorithm.

\bibitem{Li2015Scalable}
J.~Li, Y.~Xia, Z.~Shan, Y.~Liu, Scalable constrained spectral clustering, IEEE
  Transactions on Knowledge \& Data Engineering 27~(2) (2015) 589--593.

\bibitem{Nie2011Spectral}
F.~Nie, Z.~Zeng, I.~W. Tsang, D.~Xu, C.~Zhang, Spectral embedded clustering: A
  framework for in-sample and out-of-sample spectral clustering, IEEE
  Transactions on Neural Networks 22~(11) (2011) 1796--1808.

\bibitem{Liu2017Spectral}
H.~Liu, J.~Wu, T.~Liu, D.~Tao, Y.~Fu, Spectral ensemble clustering via weighted
  k-means: Theoretical and practical evidence, IEEE Transactions on Knowledge
  \& Data Engineering 29~(5) (2017) 1129--1143.

\bibitem{Yu_Shi_multiclass-iccv-2003}
S.~X. Yu, J.~Shi, Multiclass spectral clustering, in: ICCV, 2003, pp. 313--319
  vol.1.

\bibitem{Nadler:2005:DMS:2976248.2976368}
B.~Nadler, S.~Lafon, R.~R. Coifman, I.~G. Kevrekidis, Diffusion maps, spectral
  clustering and eigenfunctions of fokker-planck operators, in: NIPS, NIPS'05,
  2005, pp. 955--962.

\bibitem{Kang-arxiv-unified-2017}
Z.~Kang, C.~Peng, Q.~Cheng, Z.~Xu, Unified spectral clustering with optimal
  graph, CoRR abs/1711.04258 (2017).
\newblock \href {http://arxiv.org/abs/1711.04258} {\path{arXiv:1711.04258}}.

\bibitem{Shi2000-pami-Normalized}
J.~Shi, J.~Malik, Normalized cuts and image segmentation, IEEE Transactions on
  Pattern Analysis and Machine Intelligence 22~(8) (2000) 888--905.

\bibitem{vonLuxburg2007-a-tutorial-on-spectral-clustering}
U.~von Luxburg, A tutorial on spectral clustering, Statistics and Computing
  17~(4) (2007) 395--416.

\bibitem{Dizaji-iccv2017-deepClustering}
K.~G. Dizaji, A.~Herandi, C.~Deng, W.~Cai, H.~Huang, Deep clustering via joint
  convolutional autoencoder embedding and relative entropy minimization, in:
  ICCV, 2017, pp. 5747--5756.

\bibitem{Li2017DiscriminativelyBI}
F.~Li, H.~Qiao, B.~Zhang, X.~Xi, Discriminatively boosted image clustering with
  fully convolutional auto-encoders, CoRR abs/1703.07980 (2017).

\bibitem{Shaham-iclr-2018SpectralNetSC}
U.~Shaham, K.~P. Stanton, H.~Li, B.~Nadler, R.~Basri, Y.~Kluger, Spectralnet:
  Spectral clustering using deep neural networks, CoRR abs/1801.01587 (2018).

\bibitem{Tian:2014:Learning-deep}
F.~Tian, B.~Gao, Q.~Cui, E.~Chen, T.-Y. Liu, Learning deep representations for
  graph clustering, in: AAAI, AAAI'14, 2014.

\bibitem{Xie:icml2016:Unsupevisd-deep-learning}
J.~Xie, R.~Girshick, A.~Farhadi, Unsupervised deep embedding for clustering
  analysis, in: ICML, 2016, pp. 478--487.

\bibitem{Yang2016Towards}
B.~Yang, X.~Fu, N.~D. Sidiropoulos, M.~Hong, Towards k-means-friendly spaces:
  Simultaneous deep learning and clustering (2016).

\bibitem{Yang-cvpr2016-joint}
J.~Yang, D.~Parikh, D.~Batra, Joint unsupervised learning of deep
  representations and image clusters, in: CVPR, 2016, pp. 5147--5156.

\bibitem{wangdeepgmm}
J.~Wang, J.~Jiang, An unsupervised deep learning framework via integrated
  optimization of representation learning and gmm-based modeling, in: ACCV,
  2018, pp. 249--265.

\bibitem{Aljalbout2018ClusteringWD}
E.~Aljalbout, V.~Golkov, Y.~Siddiqui, D.~Cremers, Clustering with deep
  learning: Taxonomy and new methods, CoRR abs/1801.07648 (2018).

\bibitem{chan-2015-tip-pcanet}
T.~H. Chan, K.~Jia, S.~Gao, J.~Lu, Z.~Zeng, Y.~Ma, Pcanet: A simple deep
  learning baseline for image classification?, IEEE Transactions on Image
  Processing 24~(12) (2015) 5017--5032.

\bibitem{chung-spectral-graph1997}
F.~R.~K. Chung, Spectral Graph Theory, CBMS Regional Conference Series in
  Mathematics, 1997.

\bibitem{Zelnik-Manor:2004:Self-tuning}
L.~Zelnik-Manor, P.~Perona, Self-tuning spectral clustering, in: NIPS, NIPS'04,
  2004, pp. 1601--1608.

\bibitem{zhu-cvpr2014-constructing}
X.~Zhu, C.~C. Loy, S.~Gong, Constructing robust affinity graphs for spectral
  clustering, in: CVPR, 2014, pp. 1450--1457.

\bibitem{Dhillon:2007-weighted-graph}
I.~S. Dhillon, Y.~Guan, B.~Kulis, Weighted graph cuts without eigenvectors a
  multilevel approach, IEEE Trans. Pattern Anal. Mach. Intell. 29~(11) (2007)
  1944--1957.

\bibitem{Yan:2009-fast-approximate}
D.~Yan, L.~Huang, M.~I. Jordan, Fast approximate spectral clustering, in:
  SIGKDD, KDD '09, 2009, pp. 907--916.

\bibitem{Zhang-icml-2008-improved}
K.~Zhang, I.~W. Tsang, J.~T. Kwok, Improved nystr\"{O}m low-rank approximation
  and error analysis, in: ICML, ICML '08, 2008, pp. 1232--1239.

\bibitem{Fowlkes-tpami2004-spectral-grouping}
C.~Fowlkes, S.~Belongie, F.~Chung, J.~Malik, Spectral grouping using the
  nystrom method, IEEE Transactions on Pattern Analysis and Machine
  Intelligence 26~(2) (2004) 214--225.

\bibitem{wang-icde-2015-multilevel}
L.~Wang, M.~Dong, A.~Kotov, Multi-level approximate spectral clustering, in:
  2015 IEEE International Conference on Data Mining, 2015, pp. 439--448.

\bibitem{Han-ijcnn2017-miniBatch}
Y.~Han, M.~Filippone, Mini-batch spectral clustering, in: IJCNN, 2017, pp.
  3888--3895.

\bibitem{wang-tsvt}
J.~Wang, G.~Wang, Hierarchical spatial sum product networks for action
  recognition in still images, IEEE Transactions on Circuits and Systems for
  Video Technology 28~(1) (2018) 90--100.

\bibitem{Hinton-Science2006-reducing}
G.~Hinton, R.~Salakhutdinov, Reducing the dimensionality of data with neural
  networks, Science 313~(5786) (2006) 504 -- 507.

\bibitem{Chen15a-arxiv-deeplearningwith}
G.~Chen, \href{http://arxiv.org/abs/1501.03084}{Deep learning with
  nonparametric clustering}, CoRR abs/1501.03084 (2015).
\newblock \href {http://arxiv.org/abs/1501.03084} {\path{arXiv:1501.03084}}.
\newline\urlprefix\url{http://arxiv.org/abs/1501.03084}

\bibitem{wang-icme2019}
J.~Wang, A.~Hilton, J.~Jiang, Spectral analysis network for deep representation
  learning and image clustering, in: 2019 IEEE International Conference on
  Multimedia and Expo (ICME), 2019, pp. 1540--1545.

\bibitem{Wang2016LearningCA}
J.~Wang, Z.~Wang, D.~Tao, S.~See, G.~Wang, Learning common and specific
  features for rgb-d semantic segmentation with deconvolutional networks, in:
  ECCV, 2016.

\bibitem{Coates2012Learning-deep-kmeans}
A.~Coates, A.~Y. Ng, Learning Feature Representations with K-Means, Springer
  Berlin Heidelberg, 2012.

\bibitem{Saad2011Numerical}
Y.~Saad, Numerical methods for large eigenvalue problems, Manchester University
  Press, 2011.

\bibitem{smga-Smola:2000:SGM:645529.657980}
A.~J. Smola, B.~Sch\"{o}kopf, Sparse greedy matrix approximation for machine
  learning, in: ICML, 2000, pp. 911--918.

\bibitem{mnist-lecun-1998-gradient}
Y.~Lecun, L.~Bottou, Y.~Bengio, P.~Haffner, Gradient-based learning applied to
  document recognition, Proceedings of the IEEE 86~(11) (1998) 2278--2324.

\bibitem{ar-dataset}
A.~Martinez, R.~Benavente., The ar face database, CVC Technical Report 24 (June
  1998).

\bibitem{Georghiades2002From-YaleB}
A.~S. Georghiades, P.~N. Belhumeur, D.~J. Kriegman, From few to many:
  Illumination cone models for face recognition under variable lighting and
  pose, IEEE Transactions on Pattern Analysis and Machine Intelligence 23~(6)
  (2002) 643--660.

\bibitem{cmu-pie-database}
T.~Sim, S.~Baker, M.~Bsat, The cmu pose, illumination, and expression (pie)
  database, in: Proceedings of Fifth IEEE International Conference on Automatic
  Face Gesture Recognition, 2002, pp. 46--51.

\bibitem{hipsterwars-kiapour-eccv2014}
M.~H. Kiapour, K.~Yamaguchi, A.~C. Berg, T.~L. Berg, Hipster wars: Discovering
  elements of fashion styles, in: ECCV, Springer International Publishing,
  2014, pp. 472--488.

\bibitem{Simo-fashion-cvpr2016}
E.~Simo-Serra, H.~Ishikawa, Fashion style in 128 floats: Joint ranking and
  classification using weak data for feature extraction, in: CVPR, 2016, pp.
  298--307.

\bibitem{He2016Deep}
K.~He, X.~Zhang, S.~Ren, J.~Sun, Deep residual learning for image recognition,
  in: CVPR, 2016, pp. 770--778.

\bibitem{Hsiao-iccv2017}
W.~L. Hsiao, K.~Grauman, Learning the latent "look": Unsupervised discovery of
  a style-coherent embedding from fashion images, in: ICCV, 2017, pp.
  4213--4222.

\bibitem{SimoSerraCVPR2015}
E.~Simo-Serra, S.~Fidler, F.~Moreno-Noguer, R.~Urtasun, {Neuroaesthetics in
  Fashion: Modeling the Perception of Fashionability}, in: CVPR, 2015.

\end{thebibliography}

\end{document}